\documentclass[10pt,journal,compsoc]{IEEEtran}
\usepackage{cite}
\usepackage{amsmath,amssymb,amsfonts}
\usepackage{algorithmic}
\usepackage{graphicx}
\usepackage{textcomp}
\usepackage{xcolor}

\usepackage[utf8]{inputenc} 
\usepackage[T1]{fontenc}    
\usepackage{hyperref}       
\usepackage{url}            
\usepackage{booktabs}       
\usepackage{amsfonts}       
\usepackage{nicefrac}       
\usepackage{microtype}      

\usepackage{bm}
\usepackage{comment}
\usepackage{amsmath}
\usepackage{enumerate}
\usepackage{enumitem}
\usepackage{mathrsfs}
\usepackage{makeidx}         
\usepackage{graphicx}        
\usepackage{multicol}        
\usepackage{subfigure}
\usepackage{epsfig,amssymb,latexsym}
\usepackage{psfrag}
\usepackage{algorithmic}
\usepackage{algorithm}

\usepackage{fancyhdr}
\usepackage{multirow}

\usepackage{color}

\newcommand{\mb}{\mathbb}
\newcommand{\mr}{\mathrm}
\newcommand{\mc}{\mathcal}

\newcounter{pro_counter}

\newtheorem{proposition}[pro_counter]{Proposition}

\ifCLASSINFOpdf
\else
\fi

\begin{document}
%
\title{Representative Task Self-selection for Flexible Clustered Lifelong Learning}
%
%
%
%
\author{Gan~Sun,  
        Yang~Cong,~\IEEEmembership{Senior Member,~IEEE,}
        Qianqian~Wang,
        Bineng~Zhong,
        and~Yun~Fu,~\IEEEmembership{Fellow,~IEEE}
\IEEEcompsocitemizethanks{\IEEEcompsocthanksitem G. Sun and Y. Cong is with State Key Laboratory of Robotics, Shenyang Institute of Automation, Institutes for Robotics and Intelligent Manufacturing, Chinese Academy of Sciences, Shenyang, 110016, China, Email: sungan1412@gmail.com, congyang81@gmail.com \protect\\
\vspace{-8pt}
\IEEEcompsocthanksitem G. Sun is also with University of Chinese Academy of Sciences, Beijing, China, and with the Department of Electrical and Computer Engineering, Northeastern
University, Boston, MA. 02115 USA \protect\\
\vspace{-8pt}
\IEEEcompsocthanksitem Q. Wang is with Xidian University. Xian, Shanxi, 710071, China, Email: qianqian174@foxmail.com \protect\\
\vspace{-8pt}
\IEEEcompsocthanksitem B. Zhong is with Huaqiao University, Xiamen, Fujian, 361021, China, Email:  bnzhong@hqu.edu.cn \protect\\
\vspace{-8pt}
\IEEEcompsocthanksitem Y. Fu is with the Department of Electrical and Computer Engineering
and the Khoury College of Computer and Information Sciences, Northeastern
University, Boston, MA. 02115 USA. Email: yunfu@ece.neu.edu \protect
}
\thanks{Manuscript received April 19, 2005; revised August 26, 2015.}
\thanks{(Corresponding author: Prof. Yang Cong.)}}
%
%

\markboth{Journal of \LaTeX\ Class Files,~Vol.~14, No.~8, August~2015}%
{Shell \MakeLowercase{\textit{et al.}}: Bare Demo of IEEEtran.cls for Computer Society Journals}
%

\IEEEtitleabstractindextext{%
\begin{abstract}
Consider the lifelong machine learning paradigm whose objective is to learn a sequence of tasks depending on previous experiences, e.g., knowledge library or deep network weights. However, the knowledge libraries or deep networks for most recent lifelong learning models are with prescribed size, and can degenerate the performance for both learned tasks and coming ones when facing with a new task environment (cluster). To address this challenge, we propose a novel incremental clustered lifelong learning framework with two knowledge libraries: feature learning library and model knowledge library, called \underline{F}lexible \underline{C}lustered \underline{L}ife\underline{l}ong \underline{L}earning ($\mr{FCL^3}$). Specifically, the feature learning library modeled by an autoencoder architecture maintains a set of representation common across all the observed tasks, and the model knowledge library can be self-selected by identifying and adding new representative models (clusters). When a new task arrives, our proposed $\mr{FCL^3}$ model firstly transfers knowledge from these libraries to encode the new task, i.e., effectively and selectively soft-assigning this new task to multiple representative models over feature learning library. Then, 1) the new task with a higher outlier probability will be judged as a new representative, and used to redefine both feature learning library and representative models over time; or 2) the new task with lower outlier probability will only refine the feature learning library. For model optimization, we cast this lifelong learning problem as an alternating direction minimization problem as a new task comes. Finally, we evaluate the proposed framework by analyzing several multi-task datasets, and the experimental results demonstrate that our $\mr{FCL^3}$ model can achieve better performance than most lifelong learning frameworks, even batch clustered multi-task learning models.


\end{abstract}

}

\maketitle

\IEEEdisplaynontitleabstractindextext

%
\IEEEpeerreviewmaketitle

\IEEEraisesectionheading{\section{Introduction}\label{sec:introduction}}
\IEEEPARstart{R}ecent successes of lifelong machine learning have been applied into many areas \cite{tessler2017deep,lopez2017gradient,sun2018lifelong,isele2017representations,shu2017lifelong,sun2018robust}, e.g., sentiment classification \cite{chen2015lifelong}, quadrotor control \cite{isele2016using} and reinforcement learning \cite{abel2018policy,zhan2017scalable}. Different from the standard multi-task learning methods which must jointly learn from the previously observed task data in the offline regime, lifelong learning methods explore how to establish the relationships between the observed tasks with new coming ones, and avoid losing performance among the previously encountered tasks. Generally speaking, by compressing the previous knowledge into a compact knowledge library \cite{ruvolo2013ella,ammar2014online,isele2016using} or storing the knowledge in the learned network weights \cite{kirkpatrick2017overcoming,rusu2016progressive,zenke2017continual}, the major procedure in most existing state-of-the-arts is to transfer knowledge from the current knowledge library \cite{ruvolo2013ella,ammar2014online} to learn a new coming task and congest the fresh knowledge over time; or simple-retrain the deep lifelong learning network via overcoming catastrophic forgetting \cite{kirkpatrick2017overcoming,zenke2017continual,shmelkov2017incremental,rebuffi2017icarl}.



\begin{figure}[t]
   \centering
    \includegraphics[width=250pt, height=150pt]{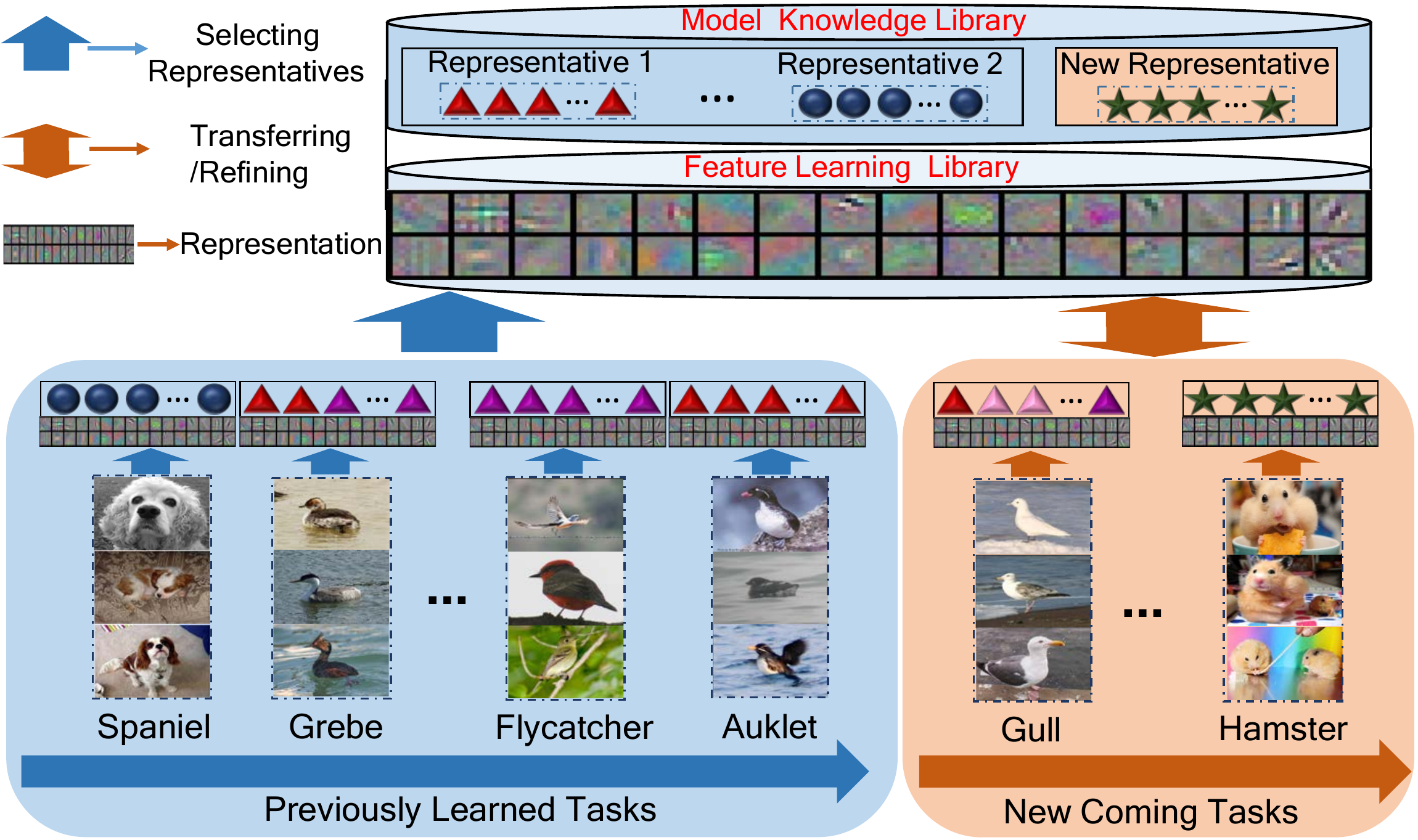}
     \vspace{-5pt}
   \caption{The demonstration of lifelong visual categorization problem. Different categories correspond to different task environments (e.g., birds have subcategories such as Auklet, Flycatcher and Grebe), where model parameters of different subcategories in the same environment share the same representative model (shape), and are assigned with different weights (colors). When a new task environment is coming, the model knowledge library should be incremental over time via adding a new representative model (e.g., rat categorization task).}
  \label{fig:categorization}
  \vspace{-15pt}
 \end{figure}

Despite the success of lifelong machine learning, the basic assumption for most existing models is that all learned tasks are drawn \emph{i.i.d.} from a distribution with compact support (i.e., task environment introduced in \cite{Bakker:2003}). This allows the learned tasks can be described with a knowledge library of latent model components, or incrementally fine-tuning the deep neural networks. However, this assumption can induce that the size of the knowledge library or neural network is prescribed, which will limit the storage capacity of knowledge library or neural network, and further result in an impractical ``lifelong learning'' system to some extent. For the future tasks, a set of encountered tasks for the lifelong learning system may be from dynamic task environments. \emph{Take the lifelong fine-grained visual categorization (FGVC) problem \cite{ristin2015categories} as an example, as illustrated in Fig.~\ref{fig:categorization}, different categories can be considered as different task environments with the situation that each environment consists of a few basis subcategories. When a sequence of subcategory tasks which are from unknown categories is input into the lifelong learning system, knowledge library/neural networks cannot transfer an effective inductive bias from learned task environments to a strange environment.} For the previous tasks, the common component/network among all task models is often invalid in many real-world problems, since negatively transferring the knowledge from unrelated tasks (i.e., unknown task environments) into the fixed knowledge library/neural network may significantly damage the performance on the previous tasks.

\indent Inspired by these aforementioned issues, in this paper, we explore how to establish an incremental lifelong machine learning system to improve knowledge transfer between earlier and later tasks. We concentrate on the lifelong learning scenarios \cite{ruvolo2013ella}, in which the accumulated experiences among previously learned tasks are stored in the knowledge library. Our approach is to extend traditional clustered multi-task learning model \cite{zhou2016flexible} into lifelong learning, called \underline{F}lexible \underline{C}lustered \underline{L}ife\underline{l}ong \underline{L}earning ($\mr{FCL^3}$), which can gradually learn the coming tasks with fixed feature learning library and an incremental model knowledge library. As shown in Fig.~\ref{fig:categorization}, the feature learning library is composed of shared representation among encountered tasks, which are learned via an autoencoder architecture; the model knowledge library consists of a set of representative models (i.e., clusters), with each representative model corresponding to an independent task environment. When a new task with unknown distribution arrives at the lifelong learning system, our proposed $\mr{FCL^3}$ maximally transfers the knowledge from the model knowledge library to represent or learn the new task model via effectively soft-assigning it into multiple representative models with sparsity weights. Next, the new task with low outlier probability can be sufficiently encoded by its representative models and further used to refine feature learning library; the new task with higher outlier probability can be used to refine feature learning library and trigger the ``birth'' of a new representative model, i.e., a new task environment. Since our proposed $\mr{FCL^3}$ framework is non-convex and NP-hard, we propose to employ the alternating direction strategy to solve this problem when a new task comes. To the end, we validate our proposed model against several knowledge library based lifelong learning models, and even multi-task learning models on one synthetic and several real-world benchmark datasets. The experiment results strongly support our $\mr{FCL^3}$ model that it can achieve similar or better performance in terms of effectiveness and efficiency.

\indent The contributions of our work are highlighted in three folds:
\vspace{-4pt}
\begin{itemize}[leftmargin=11pt]
\item Different from most existing lifelong learning models whose knowledge libraries or network structures are with prescribed size, we propose a \underline{F}lexible \underline{C}lustered \underline{L}ife\underline{l}ong \underline{L}earning framework with incremental representative models, named as $\mr{FCL^3}$, which can maximally utilize the knowledge among current knowledge libraries and dynamically increase the capacity of model knowledge library, i.e., number of representative models.
\item Two knowledge libraries are defined in our $\mr{FCL^3}$ formulation: an incremental model knowledge library is used to store a set of shared representative models; the feature learning library, which maintains a set of shared representation common across all encountered tasks is learned via an autoencoder architecture.
\item We transfer the knowledge from the most related representative models to aid the learning of new tasks: 1) the useless representative models for the new task can be filtered via sparsity constraint; 2) the new task with high outlier probability will be self-selected as a new representative, i.e., the ``birth'' of the representative. Experiments on one synthetic and several real-world datasets demonstrate the higher improvement and lower computational cost obtained by our $\mr{FCL^3}$ framework.

\end{itemize}

This paper is an extension of our conference paper \cite{gan2018clusteredlifelong}, and the new contents are as follows: 1) a unified framework which incorporates self-selecting a new representative model and learning the new task is proposed; 2) a sparse autoencoder architecture is provided to refine common representation across learning tasks, which can map model parameters of new tasks into a lower dimensional space; 3) we derive a general online formulation to alternatively update the feature learning and model knowledge libraries; 4) more competing models are used to evaluate the effectiveness of the proposed model and 5) we also conduct more sensitivity studies on the proposed model, e.g., parameter analysis, effect of the task order, and so on.

The rest of the paper is organized as follows. Section.~\ref{sec:related work} gives a brief review of some related work. Then Section.~\ref{sec:formulation} proposes our flexible clustered lifelong learning formulation. How to update the proposed framework efficiently with a new coming task via alternating direction algorithm is proposed in the subsequent Section.~\ref{sec:model_optimization}. The last two sections (i.e., Section.~\ref{sec:experiment}, and Section~\ref{sec:conclusion}) report the experimental results and conclusion of this paper.


\vspace{-5pt}
\section{Related Works}\label{sec:related work}
Since lifelong learning can be considered as an online learning framework of multi-task learning, we review the related works from two parts: \textbf{Multi-task Learning} and \textbf{Lifelong Learning}.

\subsection{Multi-task Learning}
Most state-of-the-art multi-task learning models attempt to explore what and how to share knowledge among different tasks. Since our model mainly focuses on feature-learning based MTL models and task-clustering based MTL models, in this section, we discuss several multi-task learning works which are relevant to our proposed model.


For the feature-learning based models, multi-layered feed forward neural networks \cite{Caruana:1997,Baxter:2000} propose one of the earliest models for feature transformation. In the multi-layered feed forward neural network, the common features learned from multiple tasks can be represented in the hidden layer, and the output of each task corresponds to each unit in the output layer. In addition, \cite{liao2006radial} extends the radial basis function network to MTL by determining the construction of the hidden layer. In comparison with multi-layer neural networks, \cite{Argyriou:2008} presents an algorithm to incorporate a mixed $\ell_{2,1}$-norm, and learn common sparse representations across multiple tasks. Furthermore, some multi-task feature selection models are proposed to select one subset of the original features via using some sparsity-inducing regularizers, e.g., $\ell_{1}+\ell_{1,\infty}$-norm (i.e. dirty model \cite{Jalali:2010}) which is used to leverage the shared features common across tasks, $\ell_{2,1}$-norm \cite{Liu:2009} which can capture the shared features via inducing a row-sparse matrix. More introduction can be found in surveys \cite{zhang2017survey}.

For the task-clustering based MTL models~\cite{Zhou:2011,Jacob:2008,Bakker:2003,Xu:2015}, the main idea is that all the task can be partitioned into several clusters, and the task parameters within each cluster are either sharing a common probabilistic prior or close to each other in distance metric. A benefit of this model is its robustness against outlier tasks since they reside in independent clusters that do not affect other tasks. However, these models might fail to take benefit of negatively correlated tasks because they can just put these in different clusters. Furthermore, \cite{zhou2016flexible} clusters multiple tasks by identifying a set of representative tasks, and an arbitrary task can be described by multiple representative tasks. The objective function of this method is:
\begin{equation}
\begin{aligned}\label{eq:fcmtl}
\min_{W,b,}&\;\; \mc{L}(W)+\gamma\left\|W\right\|_F^2+\lambda\sum_{i=1}^m\sum_{k=1}^mZ_{ik}\left\|w_i-w_k\right\|_2^2 \\ &+\mu\left\|Z\right\|_{1,2}, \\
s.t.,&\;\;   0\preceq \mr{vec}(Z)\preceq \bm{1}_{mm},\; Z^{\top}\bm{1}_{m}=1_m,
\end{aligned}
\end{equation}
where $Z$ denotes the assignment of representative tasks for all tasks. However, (i) this method which selects a subset of representative tasks in the offline regime cannot be transferred into a new task environment; (ii) discriminative features among multiple tasks are not learned during the training phase, which leads to high computational cost due to redundant features. To address these two challenges (especially the first challenge), we pay our attention to the lifelong machine learning \cite{chen2016lifelong}, which is another machine learning paradigm, and adapts the knowledge learned in the past to help future learning and problem solving.

\subsection{Lifelong Learning}
\indent For the \textbf{Lifelong Learning}, the early works~\cite{thrun1996discovering} on lifelong learning aims to transfer knowledge achieved from earlier tasks to later ones, or transfer invariance knowledge among neural networks \cite{thrun2012explanation}. For the recently-proposed traditional lifelong learning framework \cite{sun2018lifelong,sun2018active}, an efficient lifelong learning algorithm (ELLA)~\cite{ruvolo2013ella} is proposed for online lifelong learning, which is based on an existing multi-task learning formulation \cite{Abhishek:2012}. Specifically, with the assumption that model parameters of multiple related tasks share a common knowledge library, the new observed task can be efficiently learned by transferring useful knowledge from the knowledge library. Additionally, \cite{isele2016using} proposes coupled dictionary learning to utilize high-level task descriptors into lifelong learning, which can model the inter-task relationships and perform zero-shot transfer learning. \cite{pentina2015lifelong} proposes to learn an inductive bias in form of a transfer procedure, which is the earlier research about lifelong learning with non-stationary data distribution, i.e., observed tasks in lifelong learning system may not form an i.i.d sample. However, none of these works consider to extend clustering structure to lifelong learning system while adding the cluster centers adaptively as new tasks arrive at the system.

Different from traditional lifelong machine learning models, deep learning framework are also adopted into lifelong learning \cite{rannen2017encoder}. Specifically, deep lifelong learning can be straightforwardly achieved by simple retraining the original neural network architecture. For instance, \cite{li2016learning} proposes a method to learning convolutional neural network without forgetting, which can retain performance on original tasks through knowledge distillation \cite{hinton2015distilling}, and train the network using only the data of the new task. To dynamically decide its network capacity as a sequence of tasks come, \cite{yoon2018lifelong} proposes a Dynamically Expandable Network (DEN) for lifelong learning. Intuitively, the difference between our $\mr{FCL^3}$ model and DEN is that our $\mr{FCL^3}$ model can be beneficial for early learned tasks via reverse transferring, whereas DEN focuses on how to learn the new coming ones via dynamically network.

\section{Flexible Clustered Lifelong Learning Framework}\label{sec:formulation}
\textbf{Problem Setup:} suppose a general lifelong machine learning system faces a sequence of supervised learning tasks: $\mc{Z}^1, \mc{Z}^2,...,\mc{Z}^m$, where each task $\mc{Z}^t=(f^t,X^t,Y^t)$ can be defined by a mapping $f^t: X^t\rightarrow Y^t$, $X^t=[x_1^t,\ldots,x_{n_t}^t]\in\mb{R}^{d\times n_t}$ denotes $n_t$ data samples represented by $d$ features, and $Y^t=[y_1^t,\ldots,y_{n_t}^t]\in\mb{R}^{n_t}$ are the corresponding responses. For each task, we consider a linear mapping $f^t$ in this paper, and the mapping $f^t: X^t\rightarrow Y^t$ can be expressed as $Y^t=f(X^t;w_t)$, where $w_t\in \mb{R}^d$ denotes the task parameter of task $t$. Basically, our framework can be easily generalized to nonlinear parametric mappings. When some labeled data for some task $t$ arrive, either data for learned tasks or a new task, the lifelong learning system needs to give the predictions on the training data from previously learned tasks or new tasks.

In the next subsections, we introduce our proposed Flexible Clustered Lifelong Learning ($\mr{FCL^3}$) framework. The key insight of $\mr{FCL^3}$ is that a set of representative models (model knowledge library) over discriminative feature representation (feature learning library) can be adopted to represent or describe all learned tasks depending on the similarity among multiple tasks, and redefined over time as a series of new tasks arrive. Therefore, in the following, we firstly describe how to define the feature learning library with an autoencoder architecture when learning a new task. Then we provide how to represent or describe the new task model via representative models and self-select the new representative model via an outlier detection strategy, followed by how to solve the proposed $\mr{FCL^3}$ framework via the alternating direction method.



\begin{figure*}[t]
   \centering
    \includegraphics[width=510pt, height=124pt]{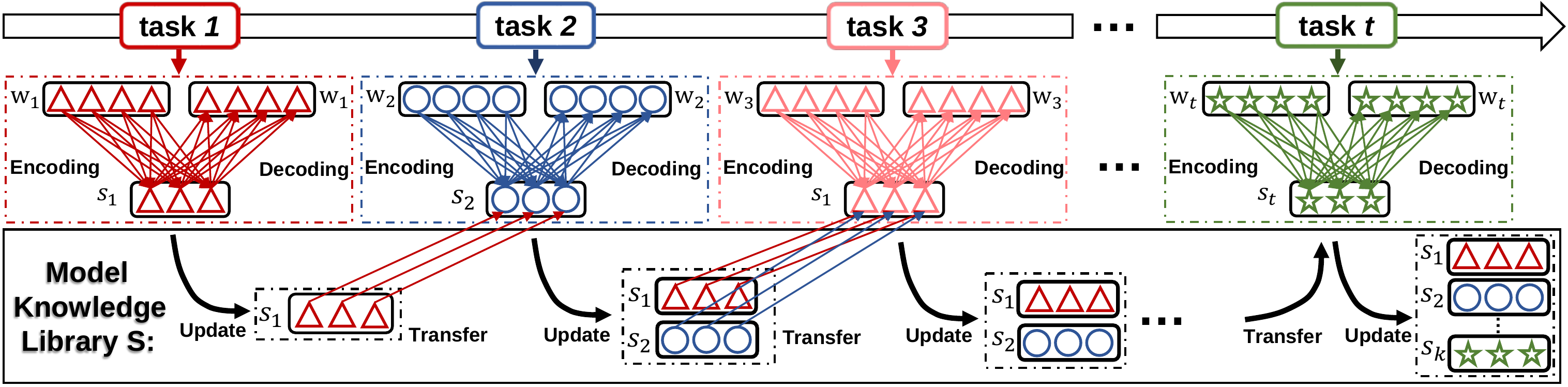}
      \vspace{-8pt}
   \caption{The demonstration of our Flexible Clustered Lifelong Learning ($\mr{FCL^3}$) framework, where the feature learning library for feature learning is achieved on the encoding and decoding phases, and $s_1$ denotes $\phi(Lw_1)$ in Eq.~\ref{eq:ClusteredLifelongLearning}. Different tasks are marked as different colors, and the tasks with similar shapes are in a same task environment, e.g., task $1$ and task $3$. The model knowledge library is initialized using the first task, and then self-selected from the following coming tasks gradually.}
   \vspace{-6pt}
  \label{fig:fcl3_model}
  \vspace{-10pt}
 \end{figure*}

\subsection{Sparse Autoencoder}
To learn the common feature representation \cite{Argyriou:2008} instead of directly using the original ones, we extend the well-known autoencoder architecture, which is composed of an input layer, a set of hidden layers, and an output layer whose attempt is to reconstruct the data of input layer. Generally, it can act as the feature learning method for multi-task learning, i.e., projecting the original task space into a lower dimensional space and back to reconstruct the original task space. In this paper, we propose to model stacked autoencoder and further preserve the task relationships via sparsity constraint. Formally, when the new task $t$ comes to the lifelong learning system, the objective formulation by incorporating a single layer generative autoencoder model can be expressed as:
\begin{equation}\label{eq:singletask}
\min\limits_{w_t,D,L}  \; \mc{L}(f(X^t; D\phi(Lw_t)),y^t)\\
 +\lambda_1\left\|\phi(Lw_t)\right\|_1,  \\
\end{equation}
where $\lambda_1\geq 0$ is the regularization parameter, $\phi$ denotes a linear or non-linear activation function such as sigmoid function, $L$ and $D$ which are called as feature learning library in this paper denote the encoding and decoding matrices, respectively. Intuitively, both $L\in\mb{R}^{p\times d}$ and $D\in\mb{R}^{d\times p}$ maintain a set of common representation among all the learned tasks, and higher order feature representation can be learnt if we stack multiple layers together. To make the formulation about $L$ and $D$ tractable, we constrain each column of $L$ and $D$ with $\left\|l_i\right\|_2\leq 1$ and $\left\|d_i\right\|_2\leq 1$, respectively.  The $\phi(Lw_t)$ in term $\left\|\phi(Lw_t)\right\|_1$ denotes the low-dimensional code vector of $w_t$, and the sparsity constraint is performed that: code vectors of learnt tasks will couple the learned tasks together when they have the same sparsity pattern, while the tasks whose code vectors are orthogonal are sure to belong to different couples.


\vspace{-3pt}
\subsection{Representative Models}
In addition to the feature learning library $\{D,L\}$ which are with fixed size, we also construct a model knowledge library $S=\{\phi(Lw_1), \ldots,\phi(Lw_K)\}$ with the situation that $\phi(Lw_k)$ is the code vector of the $k$-th learned task, and called as the $k$-th representative model. $K$ is total number of representative models in model knowledge library. When the $t$-th task (i.e., the new task) is coming, it is expected that these $K$ representative models can sufficiently capture all important task details of the $t$-th coming task. We thus present the following lifelong machine learning model by extending the problem Eq.~\eqref{eq:singletask} into multi-representative scenario:
\begin{equation}\label{eq:multi-representative}
\begin{aligned}
\min\limits_{w_t,Z^t,D,L}  &\;\; \mc{L}(f(X^t;D\phi(Lw_t)),y^t)+\lambda_1\left\|\phi(Lw_t)\right\|_1\\
 +&\lambda_2\Big(\sum_{k=1}^{K}z_{k}^t\mc{L}_k(f(X^t;D\phi(Lw_k)),y^t)+\alpha\Phi(Z^t)\Big),
 \end{aligned}
\end{equation}
where $\mc{L}_k(f(X^t;D\phi(Lw_k)),y^t)$ denotes the available knowledge transferred from the $k$-th representative model, $Z^t\in \mb{R}^{K}$ denotes the assignment of current representative models for the $t$-th task, $\lambda_1>0$, $\lambda_2>0$ and $\alpha>0$ are the corresponding regularization parameters. Each $z_k^t$ in $Z^t$ denotes that the new task $t$ could be linked to the $k$-th representative model with a probability $z_k^t$, i.e., if $z_k^t=0$, the $k$-th representative model will not contribute to the learning of the new task, and if $z_k^t=1$, it denotes that the $k$-th model will be the unique representative model of the new task. Therefore, we impose a constraint on $Z^t:\sum_{k=1}^{K}z_k^t=1$ to ensure the sum of the total probability of all representative models to be one. The $\Phi(Z^t)$ term in above formulation aims to penalize the complexity of $Z^t$, and make the problem about $Z^t$ trackable.

Additionally, in many real-world applications, it is desirable to establish relationships among these continuous tasks, i.e., similar tasks should share similar representatives (clusters) information with each other. Therefore, the corresponding number of representative models for coming task is expected to be small. Inspired by \cite{zhou2016flexible,ijcai2017-475}, we utilize sparsity inducing constraint on $Z^t$ (i.e., $\Phi(Z^t)=\left\|Z^t\right\|_1$) with the expectation that if the $k$-th element in $Z^t$ is non-zero for two coming tasks, these two tasks can be seen as belonging to the same $k$-th representative model; if at least two elements in $Z^t$ is non-zero for two coming tasks, then they are considered as in the same group; otherwise, two coming tasks whose assignment vectors are orthogonal to each other can be seen as belonging to different groups. Formally, the problem of describing the new task can be reformulated as:
\begin{equation}
\begin{aligned}\label{eq:representive_task}
\min\limits_{w_t,Z^t,D,L}  &\;\; \mc{L}(f(X^t;D\phi(Lw_t)),y^t)+\lambda_1\left\|\phi(Lw_t)\right\|_1\\
 +&\lambda_2\Big(\sum_{k=1}^{K}z_{k}^t\mc{L}_k(f(X^t;D\phi(Lw_k)),\!y^t)+\alpha\left\|Z^t\right\|_1\Big),  \\
 s.t., &\; 0\preceq \mr{vec}(Z^t)\preceq \bm{1}_{K}, (Z^t)^{\top}\bm{1}_{K}=1,
\end{aligned}
\end{equation}
where $\alpha\geq0$ is the parameter to control the sparsity of $Z^t$, $\preceq$ denotes elementwise inequality, $\bm{1}_{K}$ denotes a $K$-dimensional identity vector.

\textbf{Self-selecting New Representatives:} the basic assumption in Eq.~\eqref{eq:representive_task} is that the coming task from similar environments can be well learned using a set of representative models under feature learning library. However, when bringing into a new task environment, as illustrated in Fig.~\ref{fig:categorization}, the capability of the lifelong machine learning framework will be reserved due to the prescribed number $K$ of current representative models. Intuitively, a larger representation error will be generated when the new representative model (i.e., cluster) cannot be encoded efficiently by any representatives of model knowledge library $S$. Therefore, we propose an incremental model knowledge library $S$ to handle strange task environments, and recast variable $K$ of each task $t$ to be an incremental variable $K_t$. $K_t$ denotes the unique representative model number for the task $t$, and could be incremental as lifelong learning system encounters more task environments. To identify and self-select the new task environment, we introduce an auxiliary variable $e \in [0,1]$ connected with each coming task, whose value can indicate the probability of the $t$-th task being a new representative model. We can then rewrite Eq.~\eqref{eq:representive_task} as:
 \begin{equation}
\begin{aligned}\label{eq:representive_task1}
\min\limits_{w_t,Z^t,D,L}  & \;\;\mc{L}(f(X^t;D\phi(Lw_t)),y^t)+\lambda_1\left\|\phi(Lw_t)\right\|_1\\
 +&\lambda_2\Big(\sum_{k=1}^{K_t+1}z_{k}^t\mc{L}_k(f(X^t;D\phi(Lw_k)),\!y^t)+\alpha\left\|Z^t\right\|_1\Big),  \\
 s.t., &\; 0\preceq \mr{vec}(Z^t)\preceq \bm{1}_{K_t+1},\; (Z^t)^{\top}\bm{1}_{K_t+1}=1,
\end{aligned}
\end{equation}
where $z_{K_t+1}$ is defined as $e$ ($e\geq 0$), and $\mc{L}_{K_t+1}(f(X^t;\!D\phi(Lw_{K_t+1})),\!y^t)$ is defined as $d_0$, which gives a weight on the selection of the $t$-th task as a new representative model, i.e., the smaller the value of variable $d_0$ is, the larger the outlier probability of $e$ will be, and the more likely the $t$-th task comes from a new task environment. With this defined outlier probability $e$,
\begin{itemize}[leftmargin=11pt]
  \item when $e=0$, we have $\sum_{k=1}^{K_t}z_k=1$. The new task $t$ can be learned via the model knowledge library $S$ sufficiently, and then be regarded as from observed task environments.
  \item when $e=1$, we have $\sum_{k=1}^{K_t}z_k=0$, i.e., a higher representative error will generate after assigning it to the library $S$. The new task $t$ can thus be seen as a new representative model, i.e.., a new task environment.
\end{itemize}

\subsection{Flexible Clustered Lifelong Learning ($\mr{FCL^3}$)}
Given the labeled training data and representative models for each coming task, we introduce the flexible clustered lifelong learning framework to minimize the predictive loss over all learned tasks while encouraging the models to share the feature learning library. As illustrated in Fig.~\ref{fig:fcl3_model}, the final formulation of the proposed model can be expressed as:
\begin{equation}
\begin{aligned}\label{eq:ClusteredLifelongLearning}
\min\limits_{\{D,L\}} \!\frac{1}{T}\!\sum_{t=1}^T\!& \min\limits_{\{w_t,Z^t\}}\! \bigg\{\! \underbrace{\mc{L}(f(X^t;D\phi(Lw_t)),y^t)+\lambda_1\left\|\phi(Lw_t)\right\|_1}_{\mr{Knowledge\; of\; the}\; t-\mr{th\; Task}} \\
+&\underbrace{\lambda_2\Big(\sum_{k=1}^{K_t+1}z_{k}^t\mc{L}_k(f(X^t;D\phi(Lw_k)),\!y^t)+\alpha\left\|Z^t\right\|_1\Big)}_{\mr{Knowledge\; of\; the\; Corresponding \;Representative \;Models}}\!\bigg\} \\
 s.t., & \;\; 0\preceq \mr{vec}(Z^t)\preceq \bm{1}_{K_t+1}, (Z^t)^{\top}\bm{1}_{K_t+1}=1,
\end{aligned}
\end{equation}
where the first two terms capture the original task knowledge for the $t$-th task under feature learning library, and the last two terms show how to transfer useful knowledge from its corresponding representative models. In the next section, we present the specific optimization algorithm needed to solve the above formulation.
\vspace{-10pt}

\section{Model Optimization}\label{sec:model_optimization}
The optimization problem in Eq.~\eqref{eq:ClusteredLifelongLearning} involves the $\ell_1$-norm which is non-smooth convex and cannot obtain a close-form solution. Normally, it can be optimized via an alternating optimization strategy. However, as demonstrated in \cite{ruvolo2013ella}, standard alternating optimization strategy with all encountered tasks data is inefficient to lifelong learning framework. Therefore, in this section, when our $\mr{FCL^3}$ model receives available training data for the new task $t$, we firstly apply the Taylor expansion of $\mc{L}_k(f(X^t; D\phi(Lw_t)),y^t)$ around its corresponding representative models. Then, the detailed procedure of how to alternatively optimize each variable is then provided.
\vspace{-10pt}

\subsection{Taylor Expansion Approximation}
In order to efficiently transfer the available knowledge from representative models to learn new tasks, the second-order Taylor expansion of $\mc{L}_{k}(f(X^t; Ds_t),y^t)$ is firstly presented around its representative model $Ds_t=Ds_k$, where $s_k=\phi(Lw_k)$ denotes the $k$-th representative model in the model knowledge library $S$. Furthermore, the Taylor approximation equations can be expressed as follows:
\begin{equation}
\begin{aligned}\label{eq:taylor_expansion1}
\hspace{-10pt}\mc{L}(Ds_t)=& \mc{L}(w_t)+\langle \nabla \mc{L}(w_t), Ds_t-w_t \rangle+\frac{1}{2}\left\| Ds_t -w_t\right\|_{\Omega^t}^2 \\
\mc{L}_{1}(Ds_t)=& \mc{L}(Ds_1)\!+\!\!\langle \nabla \mc{L}(Ds_1),\! Ds_t \!-\!Ds_1 \rangle\!+\!\frac{1}{2}\!\left\| Ds_t \!-\!Ds_1\right\|_{\Omega_1^t}^2 \\
& \quad \quad\quad\quad\quad\vdots    \\
\mc{L}_{k}(Ds_t)=&\mc{L}(Ds_k)\!+\!\!\langle \nabla \mc{L}(Ds_k),\! Ds_t\! -\!Ds_k \rangle\!+\!\frac{1}{2}\!\left\|Ds_t\!-\!Ds_k\right\|_{\Omega_k^t}^2 \\
& \quad \quad\quad\quad\quad\vdots    \\
\mc{L}_{{K_t}}\!(Ds_t) \!\!=\! &\mc{L}(\!Ds_{K_t}\!)\!\!+\!\!\langle \nabla\! \mc{L}(\!Ds_{K_t}\!), \!Ds_t \!\!-\!\!Ds_{K_t}\! \rangle\!\!+\!\!\frac{1}{2}\!\left\|\!Ds_t\!\!-\!\!Ds_{K_t}\!\right\|_{\Omega_{K_t}^t}^2\!\!
\end{aligned}
\end{equation}
where $w_t$ is the single-task model parameter for the task $\mc{Z}^t$ \cite{ruvolo2013ella}, $\mc{L}(Ds_t)$ and $\mc{L}_{k}(Ds_t)$ are a simplified version of $\mc{L}(f(X^t;Ds_t),y^t)$ and $\mc{L}_k(f(X^t;Ds_t),y^t)$, respectively. $\nabla\mc{L}(w_t)$ and $\nabla\mc{L}_{k}(Ds_t)$ are the corresponding first-order gradient information around the parameter of the task $t$ and its first representative model $Ds_k$, respectively. $\Omega^t$ and $\Omega_{k}^t$ are the Hessian matrices of the loss function $\mc{L}$ evaluated at $w_t$ and $Ds_k$, and can be defined as:
\begin{equation}
\begin{aligned}\label{eq:hessian_matrix}
\Omega^t=&\frac{1}{2}\nabla^2_{Ds_t,Ds_t}\mc{L}(f(X^t;Ds_t),y^t)\mid_{Ds_t=w_t}, \\
\Omega_k^t=&\frac{1}{2}\nabla^2_{Ds_t,Ds_t}\mc{L}(f(X^t; Ds_t),y^t)\!\mid_{Ds_t=Ds_k}.
\end{aligned}
\end{equation}
After plugging these equations from Eq.~\eqref{eq:taylor_expansion1} into Eq.~\eqref{eq:ClusteredLifelongLearning} and suppressing the constant term of the above Taylor expansion, we can obtain the following formulation when each new task $t$ is coming:
\begin{equation}
\begin{aligned}\label{eq:optimization_problem}
\min\limits_{\{D,L\}}\! \frac{1}{T}\!\sum_{t=1}^T& \!\min\limits_{\{s_t,Z^t\}} \!\bigg\{\!\big(\!\left\|w_t\!-\!Ds_t\right\|_{\Omega^t}^2\!+\!\left\|s_t\!-\!\phi(Lw_t)\right\|_F^2\!\big)\!+\!\lambda_1\!\left\|s_t\right\|_1 \\
+&\lambda_2\Big(\sum_{k=1}^{K_t+1}z_{k}^t\left\|Ds_k-Ds_t\right\|_{\Omega_k^t}^2+\alpha\left\|Z^t\right\|_1\Big)\bigg\} \\
 s.t., &\;\; 0\preceq \mr{vec}(Z^t)\preceq \bm{1}_{K_t+1},\; (Z^t)^{\top}\bm{1}_{K_t+1}=1,
\end{aligned}
\end{equation}
where the second term $\!\left\|s_t\!-\!\phi(Lw_t)\right\|_{F}^2$ is the encoder stage. In order to solve the subproblem in Eq.~\eqref{eq:optimization_problem}, we invoke an alternating direction strategy to iteratively updating $\{s_t,Z^t\}$, $\{D,L\}$, and the model knowledge library $S$.

\subsection{Solving $\{s_t,Z^t\}$ with Given Feature Learning Library $\{D,L\}$}
With the fixed feature learning library $\{D,L\}$, $\{s_t,Z^t\}$ are the variables in this subproblem, and this optimization subproblem can be defined as:
\begin{equation}
\begin{aligned}\label{eq:subproblem_sz}
\min\limits_{\{s_t,Z^t\}}&\big(\left\|w_t-Ds_t\right\|_{\Omega^t}^2\!+\!\left\|s_t-\phi(Lw_t)\right\|_F^2\big)+\lambda_1\left\|s_t\right\|_1 \\
+&\lambda_2\Big(\sum_{k=1}^{K_t+1}z_{k}^t\left\|Ds_k-Ds_t\right\|_{\Omega_k^t}^2+\alpha\left\|Z^t\right\|_1\Big) \\
 s.t., &\;\; 0\preceq \mr{vec}(Z^t)\preceq \bm{1}_{K_t+1},\; (Z^t)^{\top}\bm{1}_{K_t+1}=1.
\end{aligned}
\end{equation}
\begin{proposition}
The optimization problem in Eq.~\eqref{eq:subproblem_sz} is convex with respect to $s_t$ and $Z^t$.
\end{proposition}
\begin{IEEEproof}
The proof for above proposition can be easily achieved: both the first three terms and last term in the objective function are convex respect to $s_t$ and $Z^t$. For the fourth term, a) the $k$-th element in $\sum_{k=1}^{K_t+1}z_{k}^t\left\|Ds_k-Ds_t\right\|_{\Omega_k^t}^2$ is $z_{k}^t\left\|Ds_k-Ds_t\right\|_{\Omega_k^t}^2$, the convexity can then be proved by justifying its Hessian to be positive definite (as shown in Eq.~\eqref{eq:hessian_matrix}); b) the fourth term can thus be convex as a nonnegative weighted sum of several convex functions due to the nonnegative property of $Z^t$.
\end{IEEEproof}

The problem in Eq.~\eqref{eq:subproblem_sz} cannot be optimized with respect to all the variables simultaneously, we then adopt an alternating method to solve this problem: optimizing $s_t$ by fixing $Z^t$, and then optimizing $Z^t$ by fixing $s_t$.

\subsubsection{Solving for $s_t$}
In order to update $s_t$, we fix $Z^t$ and remove the terms which are irrelevant to $s_t$. The objective function then becomes:
\begin{equation}
\begin{aligned}\label{eq:subproblem_s}
\min\limits_{s_t}&\; \big(\left\|w_t-Ds_t\right\|_{\Omega^t}^2+\left\|s_t-\phi(Lw_t)\right\|_{F}^2\big)+\lambda_1\left\|s_t\right\|_1\\
& +\!\lambda_2\!\!\sum_{k=1}^{K_t+1}z_{k}^t\left\|Ds_k\!-\!Ds_t\right\|_{\Omega_k^t}^2.
\end{aligned}
\end{equation}
Notice that the loss function and penalty term $\left\|s_t\right\|_1$ in above subproblem are convex and non-smooth convex, respectively. We thus employ the accelerate gradient method \cite{nesterov2013gradient} with a fast-global convergence rate to solve this subproblem.

\subsubsection{Solving for $Z^t$}
Next, in order to update $Z^t$, we also fix other variables except for $Z^t$ and remove the independent terms about $Z^t$. The objective problem can then become:
\begin{equation}
\begin{aligned}\label{eq:subproblem_z}
\min\limits_{Z^t}&\;\; \lambda_2 \mr{tr}(D^TZ^t)+\alpha\left\|Z^t\right\|_1 \\
 s.t., &\;\; 0\preceq \mr{vec}(Z^t)\preceq \bm{1}_{K_t+1},\; (Z^t)^{\top}\bm{1}_{K_t+1}=1,
\end{aligned}
\end{equation}
where each element $d_k$ in $D$ is defined as $\left\|Ds_k-Ds_t\right\|_{\Omega_k^t}^2$, and the last element of $D$ is $d_0$. Basically, identifying and adding new representative model for all learned tasks can be achieved via constant $d_0$. In this paper, the choice for the $d_0$ can be defined as:
\begin{equation}
d_0=-\gamma\mr{log}\Big(\frac{\min_{k}\left\|Ds_k-Ds_t\right\|_{\Omega_k^t}^2}{\sum_{k=1}^{K_t}\left\|Ds_k-Ds_t\right\|_{\Omega_k^t}^2}\Big),
\end{equation}
where $\gamma$ is a non-negative parameter. Intuitively, when the $t$-th task can be well represented or described by one representative model in $S$, the likelihood of $s_t$ being selected as a new representative model will decrease (i.e., $d_0$ should be a large value), and vice versa. Generally, an efficient optimization algorithm for Eq.~\eqref{eq:subproblem_z} is alternating direction method of multipliers (ADMM) \cite{boyd2011distributed}, and we can reformulate Eq.~\eqref{eq:subproblem_z} as:
\begin{equation}
\begin{aligned}\label{eq:subproblem_z1}
\min\limits_{Z^t,J^t}&\;\; \lambda_2 \mr{tr}(D^TJ^t)+\alpha\left\|Z^t\right\|_1+\beta\left\|Z^t-J^t\right\|_F^2 \\
 s.t., &\; 0\preceq \mr{vec}(Z^t)\preceq \bm{1}_{K_t+1}, (Z^t)^{\top}\bm{1}_{K_t+1}=1, Z^t=J^t,
\end{aligned}
\end{equation}
where $\beta>0$ is a regularizer parameter and $J^t$ denotes an auxiliary variable. Details of the ADMM procedure for Eq.~\eqref{eq:subproblem_z1} are given in \textbf{Appendix A}. Moreover, the detailed procedures of solving $\{s_t,Z^t\}$ are given in \textbf{Algorithm 1}.

\renewcommand{\algorithmicrequire}{\textbf{Input:}}
\renewcommand{\algorithmicensure}{\textbf{Output:}}
\begin{algorithm}[t]
\caption{\small Solving $\{s_t,{Z}^t\}$ via Alternating Direction Strategy}
\begin{algorithmic}[1]
\REQUIRE $\{s_k,\Omega_k^t\}_{k=1}^{K_t+1}$, $w_t\in \mb{R}^d$,$D\in\mathbb{R}^{d\times p}, L\in \mathbb{R}^{p\times d}$, $\lambda_1\geq 0,\lambda_2>0,\alpha>0$ and MAX-ITER
\ENSURE $\{s_t,{Z}^t\}$ \\
\STATE Initialize ${Z}^t=I_{K_t+1}/(K_t+1)$;
\FOR {$i=1,...,$ MAX-ITER}
\STATE Update Eq.~\eqref{eq:subproblem_s} via Accelerated Gradient Method\cite{Ji:2009};
\vspace{-10.pt}
\STATE Update ${Z}^t$ via Eq.~\eqref{eq:subproblem_z1};
\IF   {Convergence criteria satisfied}
\STATE   Save $\{s_t,{Z}^t\}$;
\STATE   Break;
\ENDIF
\ENDFOR \\
\STATE Return $\{s_t,{Z}^t\}$;
\end{algorithmic}
\end{algorithm}

\subsection{Solving Feature Learning Library $\{D,L\}$ with Obtained $\{s_t,Z^t\}$}
When we obtain the corresponding solution $\{s_t,Z^t\}$ of the $t$-th task, the subproblem about feature learning library $\{D,L\}$ can be rewritten as:
\begin{equation}
\begin{aligned}\label{eq:optimization_DL}
\min\limits_{\{D,L\}} \frac{1}{T}\sum_{t=1}^T& \bigg\{\big(\left\|w_t-Ds_t\right\|_{\Omega^t}^2+\left\|s_t-\phi(Lw_t)\right\|_{F}^2\big) \\
+&\lambda_2\sum_{k=1}^{K_t}z_{k}^t\left\|Ds_k-Ds_t\right\|_{\Omega_k^t}^2\bigg\}
\end{aligned}
\end{equation}
Next, we propose to how to redefine the feature learning library from the perspective of online dictionary learning.


\subsubsection{Solving for $D$}
To optimize variable $D$, we fix $L$ and remove the terms which are irrelevant to $D$. The objective function can be rewritten as:
\begin{equation}
\begin{aligned}\label{eq:optimization_D}
\mathop{\min}_{D} \frac{1}{T}\sum_{t=1}^T\bigg\{ \left\|w_t-Ds_t\right\|_{\Omega^t}^2+\lambda_2\sum_{k=1}^{K_t}\!z_{k}^t\left\|Ds_k-Ds_t\right\|_{\Omega_k^t}^2\!\bigg\}.
\end{aligned}
\end{equation}
In order to store the previous feature knowledge of encountered tasks, two statistical records are used in this paper:
\begin{equation}
\begin{aligned}\label{eq:statistic_variables_D}
A=& \frac{1}{T}\sum_{t=1}^T\Big(\big(s_t(s_t)^{\top}\big)\otimes \Omega_t+\lambda_2 \sum_{k=1}^{K_t}z_k^t(\nabla s_k\nabla s_k^{\top}) \otimes \Omega_k^t \Big), \\
b=& \frac{1}{T}\sum_{t=1}^T \mr{vec}\Big((s_t)\otimes \Big((w_t)^{\top}\Omega_t \Big) \Big) ,
\end{aligned}
\end{equation}
where $\nabla s_k$ in $A$ is defined as $s_k-s_t$. The global optimum for the Eq.~\eqref{eq:optimization_D} can be reached by taking the derivatives and setting them to zero, which can achieve the update equations for next $D$ via $A^{-1}b$.

\renewcommand{\algorithmicrequire}{\textbf{Input:}}
\renewcommand{\algorithmicensure}{\textbf{Output:}}
\begin{algorithm}[t]
\caption{Flexible Clustered Lifelong Machine Learning Framework}
\begin{algorithmic}[1]
\REQUIRE Training Dataset $(X^1,Y^1), \ldots, (X^t,Y^t)$, $d\geq p>0,\mu,\lambda_1,\lambda_2, \alpha, A\leftarrow \bm{0}_{d\times p,d\times p}, b\leftarrow \bm{0}_{d\times p,1},M\leftarrow \bm{0}_{p\times d},C\leftarrow \bm{0}_{d\times d},S=[]$;
\ENSURE $D,L,S$; \\
\WHILE {isMoreTrainingDataAvailable()}
\STATE  $(X^{\mr{new}},Y^{\mr{new}},t)\leftarrow \mr{getTrainingData()}$;
\IF {isNewTask($t$) }
\STATE $T \leftarrow T+1$; $X^{t}\leftarrow X^{\mr{new}}$, $Y^{t}\leftarrow Y^{\mr{new}}$;
\ELSE
\STATE $X^t\leftarrow [ X^t, X^{\mr{new}}]$, $Y^t\leftarrow [Y^t, Y^{\mr{new}}];$
\ENDIF
\IF {istheFirstTask}
\STATE  Compute $(\{w_1\},\Omega^1)$ from CollectedData $(f^1, X^1, Y^1)$;
\STATE Initialize $D$ and $L$ in the first coming task;
\STATE  Compute $s_1$ via Eq.~\eqref{eq:subproblem_s};
\ELSE
\STATE  Compute $(\{w_t\},\Omega^t,\{\Omega_k^t\}_{k=1}^{K_t})$ from CollectedData $(f^t, X^t, Y^t)$;
\STATE  Compute $\{s_t,{Z^t}\}$ via \textbf{Algorithm 1};
\ENDIF
\STATE  Update library $D$ via Eq.~\eqref{eq:optimization_D};
\STATE  Update library $L$ via Eq.~\eqref{eq:optimization_L};
\IF {isNewRepresentativeModel($t$) }
\STATE Update model knowledge library $S=[S,s_t]$;
\ENDIF
\ENDWHILE \\
\STATE Return $D,L,S$;
\end{algorithmic}
\end{algorithm}

\subsubsection{Solving for $L$}
To optimize variable $L$, we fix $D$ and remove the terms which are irrelevant to $L$. The objective function corresponding to $L$ can be expressed as its equivalent form:
\begin{equation}
\begin{aligned}\label{eq:optimization_L}
&\mathop{\min}\limits_{L} \frac{1}{T}\sum_{t=1}^T \left\|s_t-\phi(Lw_t)\right\|_{F}^2\equiv \left\|\phi^{-1}(s_t)-Lw_t\right\|_{F}^2. \\
\end{aligned}
\end{equation}
where this strategy has been successfully used to solve autoencoders \cite{gogna2017semi} problem. To store the previous task knowledge, we also define two statistical records:
\begin{equation}\label{eq:statistic_variables_L}
  M_t=M_{t-1}+\phi^{-1}(s_t)w_t^{\top}, C_t=C_{t-1}+w_tw_t^{\top},
\end{equation}
where $M_{t-1}=\sum_{i=1}^{t-1}\phi^{-1}(s_i)w_i^{\top}$, $C_{t-1}=\sum_{i=1}^{t-1}w_iw_i^{\top}$, and the knowledge of new task is $\phi^{-1}(s_t)w_t^{\top}$ and $w_tw_t^{\top}$. By setting the derivative of Eq.~\eqref{eq:optimization_L} to zero, each row of next $L$ can be solved via using the linear system: $M_t^i=L(i,:)C_t^i$.

Finally, the optimization procedure of our $\mr{FCL^3}$ framework is summarized in \textbf{Algorithm 2}.

\subsection{Convergence  Analysis}
This subsection presents the corresponding theoretical guarantee that the performances of early learned task can converge to a stationary point as encountering more new tasks. As first glance, we should provide the convergence analysis of both $D$ and $L$ since the $s_t$'s in the learned tasks are fixed. Therefore, we fist give the convergence analysis of variable $D$, followed by the convergence analysis of $L$.
\subsubsection{Convergence Analysis of $D$}
To begin with this convergence analysis, we first give the following proposition about the convergence rate of $D$:
\begin{proposition}
 $D$ converges asymptotically to a stationary point as the number of the learned tasks $T$ increases, and the convergence rate is $O(\frac{1}{T})$.
\end{proposition}
\begin{IEEEproof}
  To begin with this proof, we firstly define the following minimization problem about $D$:
  \begin{equation}\label{eq:minimization_D}
  \mc{H}_T(D)=\frac{1}{T}\sum_{t=1}^Th(D,w_t,s_t,\Omega^t,\{s_k,\Omega_k^t\}_{k=1}^{K_t+1},Z^t),
  \end{equation}
  where
  \begin{equation}
  \begin{aligned}\label{eq:function_h}
  &h(D,w_t,s_t,\Omega^t,\{s_k,\Omega_k^t\}_{k=1}^{K_t+1},Z^t) \\
    =&\left\|w_t-Ds_t\right\|_{\Omega^t}^2+\lambda_2\sum_{k=1}^{K_t+1}z_{k}^t\left\|Ds_k-Ds_t\right\|_{\Omega_k^t}^2.
    \end{aligned}
  \end{equation}
From the equation above, we can notice that $\mc{H}_T(D)-\mc{H}_{T-1}(D)$ is Lipschitz-continuous with constant $O(\frac{1}{T})$:
\begin{equation}\label{eq:lipschitz_constant}
\begin{aligned}
  & \mc{H}_T(D)-\mc{H}_{T-1}(D) \\
  =&\frac{1}{T}h(D,w_T,s_T,\Omega^T,\{s_k,\Omega_k^T\}_{k=1}^{K_T+1},Z^T)\\
  &+\frac{1}{T}\sum_{t=1}^{T-1}h(D,w_t,s_t,\Omega^t,\{s_k,\Omega_k^t\}_{k=1}^{K_t+1},Z^t)\\
  &-\frac{1}{T-1}\sum_{t=1}^{T-1}h(D,w_t,s_t,\Omega^t,\{s_k,\Omega_k^t\}_{k=1}^{K_t+1},Z^t) \\
  =&\frac{1}{T}h(D,w_T,s_T,\Omega^T,\{s_k,\Omega_k^T\}_{k=1}^{K_T+1},Z^T)\\
   &-\frac{1}{T(T-1)}\sum_{t=1}^{T-1}h(D,w_t,s_t,\Omega^t,\{s_k,\Omega_k^t\}_{k=1}^{K_t+1},Z^t).
  \end{aligned}
\end{equation}
Intuitively, $\mc{H}_T(D)-\mc{H}_{T-1}(D)$ equals to the difference of two components: the first one is $h$ divided by $T$, and the second one is an average over $T-1$ and normalized by $T$ (i.e., this term can obtain Lipschitz constant no greater than the largest Lipschitz constant of the $h$'s being averaged. From the Eq.~\eqref{eq:function_h}, we can easily find that the $h$ is Lipschitz-continuous with $O(1)$ since Eq.~\eqref{eq:function_h} is a quadratic function over a compact region. Therefore, $\mc{H}_T(D)-\mc{H}_{T-1}(D)$ is Lipschitz-continuous with constant $O(\frac{1}{T})$, which is set as $\xi_T$ in this paper. We then have:
\begin{equation}
\begin{aligned}\label{eq:convergence_D1}
  & \mc{H}_{T-1}(D_{T})-\mc{H}_{T-1}(D_{T-1}) \\
=& \mc{H}_{T-1}(D_{T})-\mc{H}_{T}(D_{T})+\mc{H}_{T}(D_{T})-\mc{H}_{T}(D_{T-1})\\
&+\mc{H}_{T}(D_{T-1})-\mc{H}_{T-1}(D_{T-1}) \\
\leq & \mc{H}_{T-1}(D_{T})-\mc{H}_{T}(D_{T})+\mc{H}_{T}(D_{T-1})-\mc{H}_{T-1}(D_{T-1}) \\
\leq & \xi_T \left\|D_T-D_{T-1}\right\|_F.
  \end{aligned}
\end{equation}
where $D_{T-1}$ and $D_T$ are the variables when the $T-1$-th and $T$-th tasks comes. Additionally, we assume that the smallest eigenvalue of the semi-definite positive Hessian matrix defined in Eq.~\eqref{eq:function_h} is greater than or equivalent to a non-zero constant
$\hat{\xi}$, where this hypothesis is in practice verified experimentally after a few iterations of our proposed algorithm, as shown in Fig.~\ref{fig:regression_learnedtask} and Fig.~\ref{fig:classification_learnedtask}. Then we can have:
  \begin{equation}
\begin{aligned}\label{eq:convergence_D2}
  \mc{H}_{T-1}(D_{T})-\mc{H}_{T-1}(D_{T-1}) \geq& \hat{\xi}  \left\|D_T-D_{T-1}\right\|_F^2.
  \end{aligned}
\end{equation}
To sum up, the convergence rate of $D$ can achieve by combining Eq.~\eqref{eq:convergence_D1} and Eq.~\eqref{eq:convergence_D2}:
   \begin{equation}
\begin{aligned}\label{eq:convergence_D2}
  \left\|D_T-D_{T-1}\right\|_F\leq \frac{\xi_T}{\hat{\xi} }=O(\frac{1}{T}).
  \end{aligned}
\end{equation}
\end{IEEEproof}
\subsubsection{Convergence Analysis of $L$}
The convergence rate of variable $L$ can be provided in the following proposition.
\begin{proposition}
 $L$ converges asymptotically to a stationary point as the number of the learned tasks $T$ increases, and the convergence rate is $O(\frac{1}{T})$.
\end{proposition}
\begin{IEEEproof}
A proof for the above proposition can be easily done by extending the proof of \emph{Proposition 2}.
\end{IEEEproof}

\subsection{Computational Complexity}
For our model, the main computational cost of learning a new task involves two subproblems: one optimization problem lies in Eq.~\eqref{eq:subproblem_sz}, the other one is in the Eq.~\eqref{eq:optimization_DL}. More specifically, each update begins by computing the single-task model parameter $w_t$, which is a cost of $O(\xi(d,n_t))$, where $\xi(\cdot)$ depends on the single-task learner \cite{ruvolo2013ella}, and $n_t$ denotes the number of data samples for the task $\mc{Z}^t$. Then:
\begin{itemize}[leftmargin=11pt]
  \item For the problem in Eq.~\eqref{eq:subproblem_sz}, the cost of updating $s_t$ is $O(p^2d^3)$ \cite{ruvolo2013ella}, where $p$ is the size of library $D$. Then the computation of the $Z^t$ consists of three steps of \textbf{Algorithm 3}: minimizing the Lagrangian function of Eq.~\eqref{eq:subproblem_z1} with respect to $Z^t$ can be done in $O(K_t+1)$, since we can perform the minimization in the step 3 of \textbf{Algorithm 3} via $K_t+1$ independent smaller optimization programs over the $K_t+1$ elements of $Z^t$; minimizing the Lagrangian function of Eq.~\eqref{eq:subproblem_z1} with respect to $J^t$ can be done using the algorithm in \cite{duchi2008efficient} with $O((K_t+1)\mr{log}(K_t+1))$; the update on $U$ has $O(K_t+1)$ computational time and can be performed, respectively. Therefore, the overall complexity of solving Eq.~\eqref{eq:subproblem_sz} is $O(p^2d^3+\mr{max}(K_t+1,(K_t+1)\mr{log}(K_t+1)))$.
  \item For the problem in Eq.~\eqref{eq:optimization_DL}, the cost of updating $D$ is $O(p^2d^3)$ \cite{ruvolo2013ella}. Next, the updating of $L$ involves a $d\times d$ matrix, and the computational cost is $O(pd^2+d^3)$.
\end{itemize}
Finally, when a new task is coming, the overall computational complexity of our proposed model is $O\big(\xi(d,n_t)+p^2d^3+\mr{max}(K_t+1,(K_t+1)\mr{log}(K_t+1))\big)$.

\section{Experiments}\label{sec:experiment}
In this section, we present the experimental results of our $\mr{FCL^3}$ model compared with the state-of-the-art lifelong machine learning and clustered multi-task learning models. In general, several adopted competing models are firstly introduced. Then we provide several experimental results about the effectiveness and efficiency of our model. Finally, we also discuss the capability of our model on discovering the new task environments.

\subsection{Comparison Models and Measurements}
To validate the effectiveness of our proposed $\mr{FCL^3}$ model, in this experiment, we compare our proposed model with the following baseline, clustered multi-task learning and lifelong learning models:
\begin{itemize}[leftmargin=11pt]
  \item Single Task Learning (STL): a baseline model, in which multiple input tasks are learned in an independent way;
  \item Clustered Multi-task Learning (CMTL)~\cite{Zhou:2011}: this model assumes that multiple tasks can be partitioned into a set of groups, where the similar tasks are in the same group, and the prior information about the group number is unknown;
  \item Disjoint Group Multi-task Learning (DG$\_$MTL)~\cite{Kang:2011}: this model assumes that tasks are either related or unrelated, and task groups are disjoint. The objective function is to minimize the square of trace-norm of each group's weight sub-matrix;
  \item Flexible Clustered Multi-task Learning (FCMTL)~\cite{zhou2016flexible}: an arbitrary task in FCMTL is allowed to be described by multiple representative tasks, and each task can be assigned into different clusters with different weights;
  \item Asymmetric Multi-task Learning (AMTL)~\cite{lee2016asymmetric}: this model aims to minimizes the influence of negative transfer by allowing asymmetric transfer between the tasks based on task relatedness as well as the amount of individual task losses;
  \item Curriculum Learning (CL)~\cite{pentina2015curriculum}: this model for multiple tasks proposes to firstly establish best task order, and then learn subsequent tasks based on this order;
  \item Neurogenetic Online Dictionary Learning (NODL) \cite{Garg2017Neurogenesis}: this online dictionary learning model builds a dictionary in non-stationary environments, where dictionary elements are added via continuous birth and death.
  \item ELLA-Rand~\cite{ruvolo2013ella}: an efficient lifelong machine learning model, in which new tasks arrive in a random manner;
  \item ELLA-Info~\cite{ruvolo2013active}: an active task selection model based on ELLA, where the next selected task should obtain the max expected information gain on the knowledge library of ELLA.
  \item Clustered Lifelong Learning (CL3)~\cite{gan2018clusteredlifelong}: our previous conference work, which formulates identifying new representative and learning the coming task into two different objective functions.
\end{itemize}
All the models are performed in MATLAB, and all the parameters of models are in range $\{10^{-3}\times i\}_{i=1}^{10}\cup \{10^{-2}\times i\}_{i=2}^{10}\cup \{10^{-1}\times i\}_{i=2}^{10}\cup \{2\times i\}_{i=1}^{10}\cup \{40\times i\}_{i=1}^{20}\cup \{1000\times i\}_{i=1}^{10}$. Moreover, all the used optimization algorithms are terminated depending on the criteria: a) the objective value change in two consecutive iterations is smaller than $10^{-5}$; b) the iteration number is greater than $10^5$. For the evaluation, we adopt the AUC (area under curve) and RMSE (root mean squared error) for the classification and regression problems, respectively. The bigger the AUC value is, the better the classification performance of the corresponding model will be; the smaller the RMSE value is, the better the regression performance of the corresponding model will be.



\subsection{Experimental Datasets}
In this subsection, six benchmark datasets are adopted for our experiments, including one synthetic dataset and five real-world datasets:


\noindent \textbf{Disjoint} dataset: this constructed synthetic dataset is composed of 3 clusters, where each cluster contains 10 regression tasks and each task is represented by a 40-dimensional weight vector. More specifically, each cluster center $w^c$ for the $c$-cluster is sampled from $\mathcal{N}(0,900)$. To construct the situation that different cluster $w^c$'s disjoint to the other cluster centers, the model parameters for a specific cluster are nonzero only for corresponding tasks, and are zero for all other tasks. Each task-specific component $w_i^c$ in the $c$-cluster is established as follows: (1) sample non-zero values from $\mathcal{N}(0,16)$ for the first 20 elements; (2) sample non-zero values from $\mathcal{N}(0,16)$ for the locations corresponding to the non-zero elements of $w^c$. Then the $\hat{w}_i^c$ which is the $i$-th task from the $c$-th cluster is the sum of its cluster center $w_i^c$ and a task-specific component $w_i^c$, i.e., $\hat{w}_i^c=w^c+w_i^c$.

\noindent\textbf{School} dataset\footnote{http://cvn.ecp.fr/personnel/andreas/code/mtl/index.html}: School data is a more popular dataset in multi-task learning field. This dataset consists of examination records of 15362 students (samples) from 139 secondary schools, which records their examination scores in three years (1985-87). Each sample can be described by 27 binary attributes, which includes gender, year, examination score, etc., plus 1 bias attribute. The response (target) is the examination score. Moreover, the total number of all task is 139 with each school is treated as a task.

\begin{table*}[t]
\caption{Statistics details of the used benchmark datasets in our experiment.}
\vspace{-6.0pt}
\centering
\scalebox{1.00}{
\begin{tabular}{|c|c|c|c|c|}
\hline
{Dataset Name}&$\#$ Task Numbers &$\#$ Samples & $\#$ Feature Dimension& $\#$ Problem Type \\
   \hline\hline
  \centering \textbf{Disjoint} & 30 & 50& 40 & Synthetic Regression \\
   \hline
  \centering \textbf{School} & 139 & 23 $\sim$ 251 & 28 & Score Regression \\
   \hline
  \centering \textbf{Parkinson-Motor} & 42 & 101 $\sim$ 168 & 16 & Score Regression \\
   \hline\hline
  \centering \textbf{Smart Meter} & 16 &1781 $\sim$ 4232 & 81 & User Classification \\
   \hline
  \centering \textbf{Landmine} & 29 & 445 $\sim$ 690 & 9 &   Classification \\
    \hline
  \centering \textbf{Caltech-Birds} &24 & 208 $\sim$ 240 & 128 & Image Classification \\
   \hline
\end{tabular}
}
\label{table:dataset}
\vspace{-10pt}
\end{table*}

\begin{figure}[t]
   \centering
    \includegraphics[width=250pt, height=106pt]{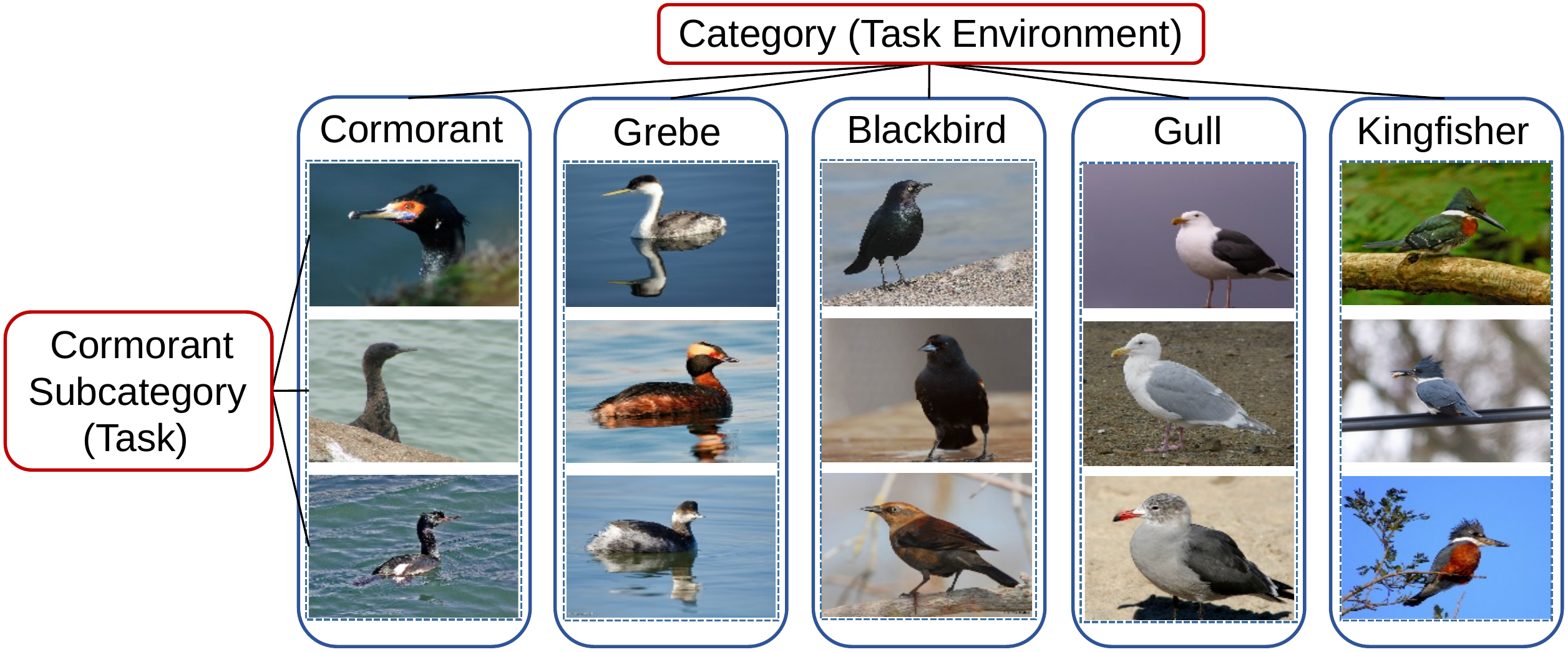}
   \caption{Example images of Caltech-Birds dataset, where each image corresponds to one classification task, and each category classification problem can be considered as one task environment.}
  \label{fig:example_image}
  \vspace{-10pt}
 \end{figure}

\begin{table*}[t]
\caption{Comparisons between our proposed $\mr{FCL^3}$ model and state-of-the-arts in terms of RMSE or AUC among six datasets: mean and standard errors averaged over ten random runs. Models with the best performance are bolded.}
\vspace{-5.0pt}
\centering
\scalebox{0.85}{
\begin{tabular}{|c|c|ccc|c|c|ccc|c|c|c}
\hline
\multicolumn{1}{|c|}{} &\multicolumn{5}{c|}{Regression Datasets} & \multicolumn{5}{c|}{Classification Datasets} \\ \hline
 {Models}&Evaluation&Disjoint & Parkinson & School  &Average &Evaluation& Landmine & SmartMeter &Caltech-Bird& Average  \\
 \hline
 STL & RMSE &0.957$\pm$0.04 &  2.346$\pm$0.15 &10.669$\pm$0.11& 4.657$\pm$0.10 &AUC($\%$)& 73.857$\pm$0.60 & 67.850$\pm$0.30& 91.455$\pm$0.78 & 77.721$\pm$0.56  \\ \hline

 CMTL\cite{Zhou:2011}& RMSE &0.780$\pm$0.01  &2.009$\pm$0.02 &10.332$\pm$0.05&4.374$\pm$0.03&AUC($\%$)&76.725$\pm$0.89&71.625$\pm$0.28& 95.515$\pm$0.63 &80.979$\pm$0.60  \\ \hline

 DG$\_$MTL\cite{Kang:2011} &RMSE &0.756$\pm$0.01  &2.073$\pm$0.01&10.149$\pm$0.07&4.326$\pm$0.03&AUC($\%$)&76.727$\pm$0.79 &  71.845$\pm$0.30& 94.295$\pm$0.37 &  80.955$\pm$0.48  \\ \hline

 FCMTL\cite{zhou2016flexible} &RMSE & 0.769$\pm$0.01 & 2.092$\pm$0.01&10.152$\pm$0.03&4.337$\pm$0.02&AUC($\%$)&77.425$\pm$0.85& 71.497$\pm$0.21& \textbf{95.523$\pm$0.27} & 81.482$\pm$0.23    \\ \hline

 AMTL \cite{lee2016asymmetric} &RMSE &0.861$\pm$0.01 &2.027$\pm$0.03 &10.614$\pm$0.06&4.507$\pm$0.03&AUC($\%$)&77.422$\pm$0.66&69.989$\pm$0.23 &94.858$\pm$0.58&80.756$\pm$0.49   \\ \hline

 CL\cite{pentina2015curriculum} &RMSE &NaN &NaN &NaN &NaN&AUC($\%$) &74.100$\pm$1.34 &68.481$\pm$0.60 &93.070$\pm$0.71&78.550$\pm$0.88 \\ \hline
NODL\cite{Garg2017Neurogenesis} &RMSE& 0.934$\pm$0.01&2.246$\pm$0.13&10.643$\pm$0.08 &4.617$\pm$0.07&AUC($\%$) &74.408$\pm$1.61&68.958$\pm$1.62 &91.345$\pm$1.19 &78.237$\pm$1.47 \\ \hline
 ELLA-Rand\cite{ruvolo2013ella} &RMSE  &0.767$\pm$0.03 &1.996$\pm$0.01& 10.165$\pm$0.08 & 4.309$\pm$0.04&AUC($\%$) & 77.429$\pm$0.79& 71.467$\pm$0.30 &  95.004$\pm$0.41 & 81.300$\pm$0.50  \\  \hline

 ELLA-Info\cite{ruvolo2013active}  &RMSE& 0.737$\pm$0.01&1.988$\pm$0.03&10.172$\pm$0.06 &4.299$\pm$0.03&AUC($\%$) &77.980$\pm$0.38&71.718$\pm$0.19 &95.273$\pm$0.53 &81.689$\pm$0.48 \\ \hline

 CL3-Rand\cite{gan2018clusteredlifelong}&RMSE& 0.714$\pm$0.01 & 1.982$\pm$0.03 & 9.995$\pm$0.02 & 4.230$\pm$0.03&AUC($\%$)&78.904$\pm$0.47&72.238$\pm$0.10 &95.327$\pm$0.38&82.156$\pm$0.32\\ \hline

 Ours-Rand  &RMSE& \textbf{0.709$\pm$0.01} & \textbf{1.974$\pm$0.03}& \textbf{9.985$\pm$0.03} & \textbf{4.223$\pm$0.02}&AUC($\%$)&\textbf{79.022$\pm$0.42}&\textbf{72.357$\pm$0.11} &95.427$\pm$0.48&\textbf{82.268$\pm$0.36}\\ \hline

 \end{tabular}
}
\label{table:regression&classification_result}
\end{table*}

\begin{figure*}[htbp]
\center
 \hspace{-4.4mm}
    \subfigure[Ground Truth]{
        \begin{minipage}[a]{0.24\textwidth}
          \centering
    \includegraphics[width =95pt ,height =85pt]{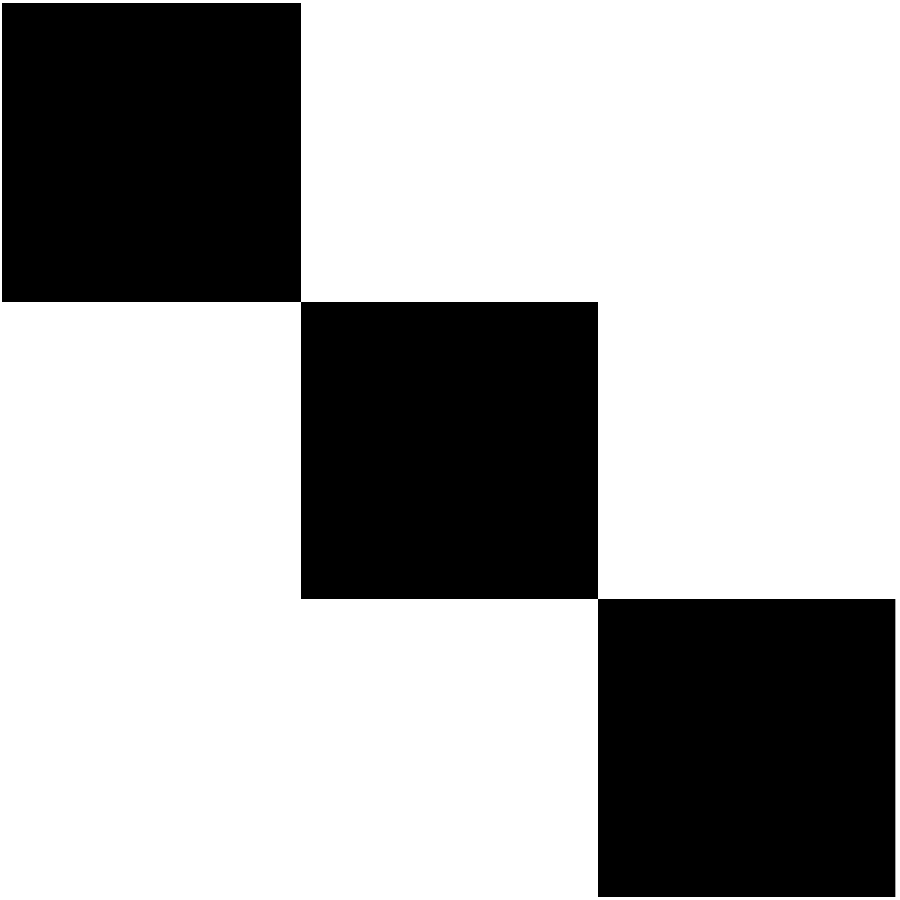}
         \end{minipage}}
         \hspace{-11.4mm}
       \subfigure[STL]{
        \begin{minipage}[a]{0.24\textwidth}
          \centering
    \includegraphics[width =95pt ,height =85pt]{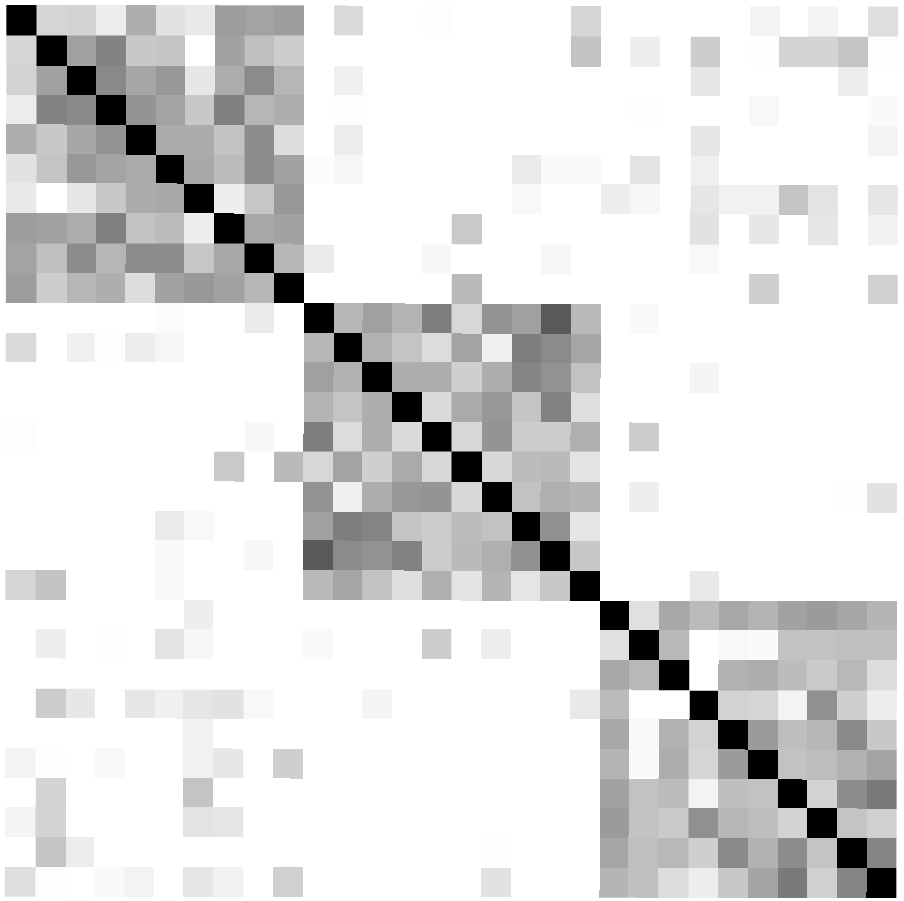}
        \end{minipage}}
        \hspace{-11.4mm}
       \subfigure[CMTL]{
        \begin{minipage}[a]{0.24\textwidth}
          \centering
    \includegraphics[width =95pt ,height =85pt]{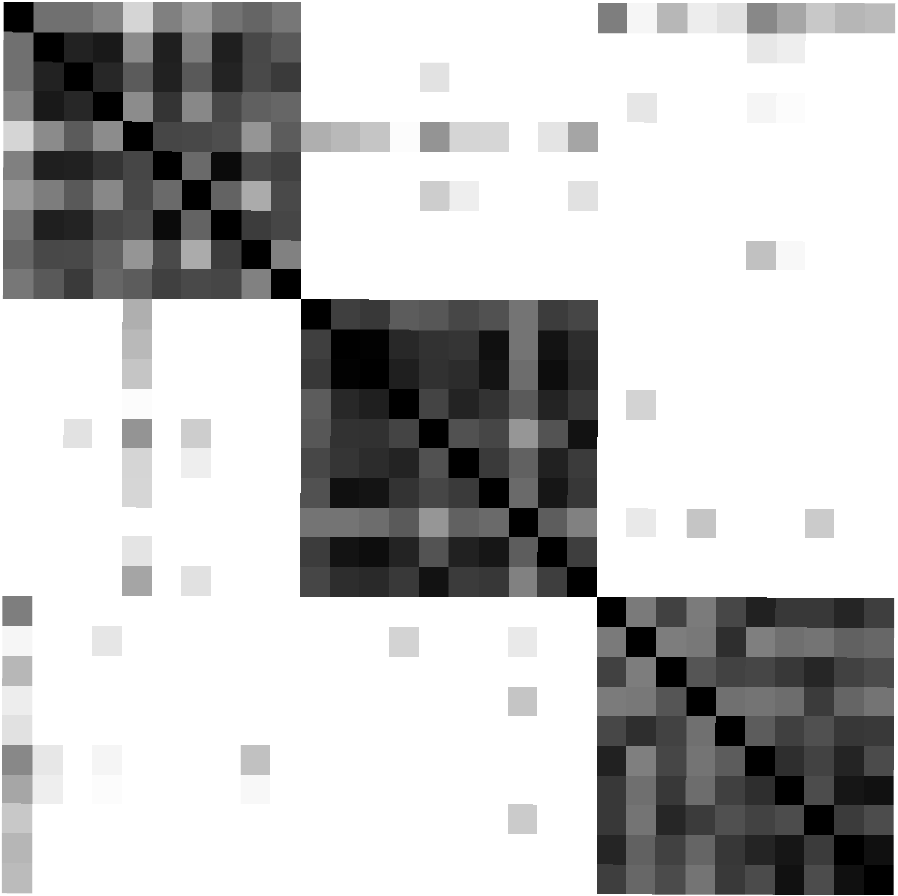}
        \end{minipage}}
       \hspace{-11.4mm}
      \subfigure[DG$\_$MTL]{
        \begin{minipage}[a]{0.24\textwidth}
          \centering
    \includegraphics[width =95pt ,height =85pt]{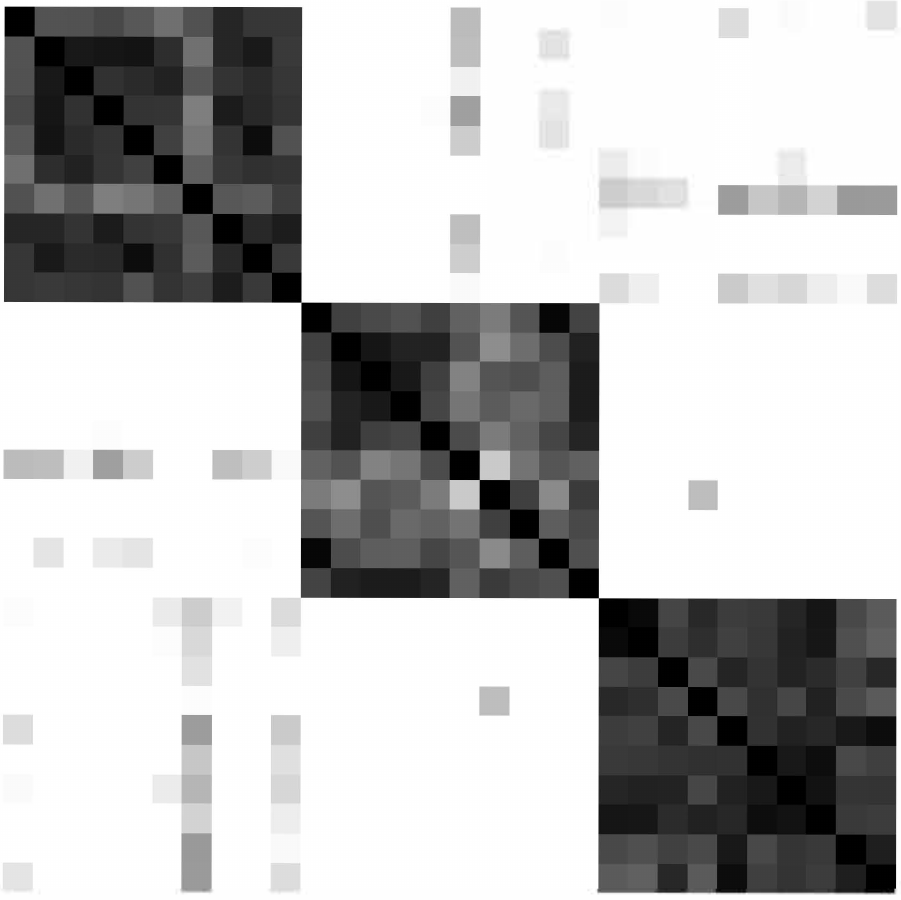}
        \end{minipage}}
      \hspace{-11.4mm}
      \subfigure[AMTL]{
        \begin{minipage}[a]{0.24\textwidth}
          \centering
    \includegraphics[width =95pt ,height =85pt]{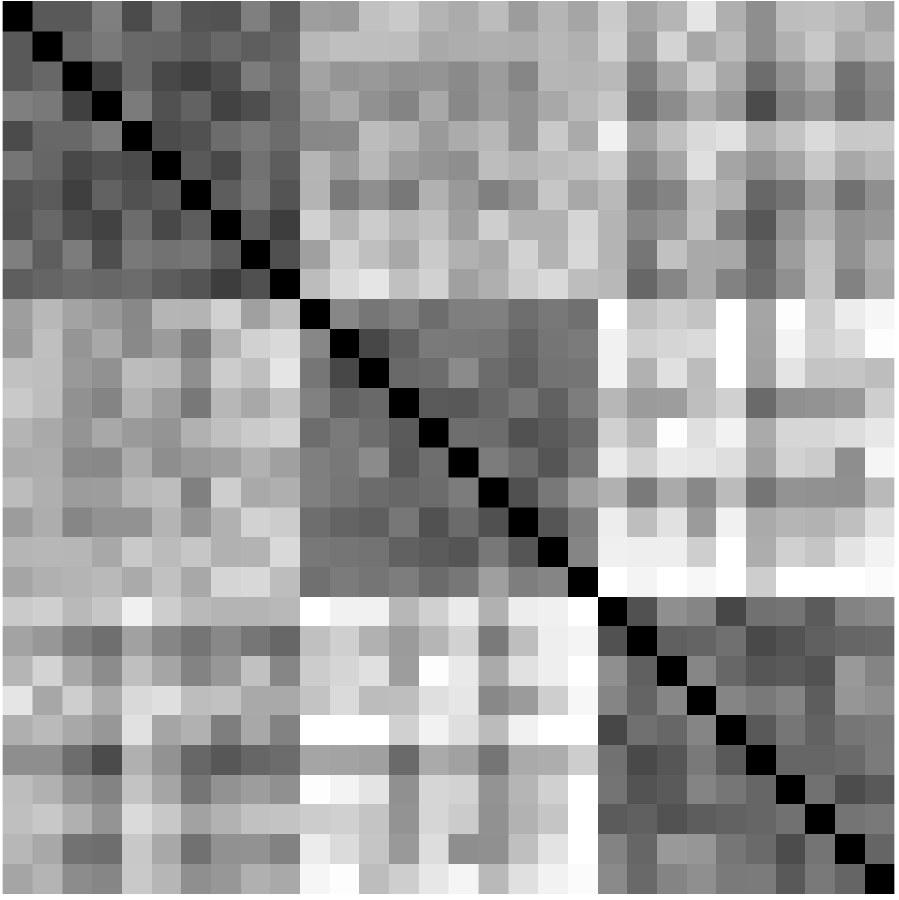}
        \end{minipage}}
        \hspace{0.4mm}  \\
        \hspace{-9.4mm}
     \subfigure[FCMTL]{
        \begin{minipage}[a]{0.24\textwidth}
          \centering
    \includegraphics[width =95pt ,height =85pt]{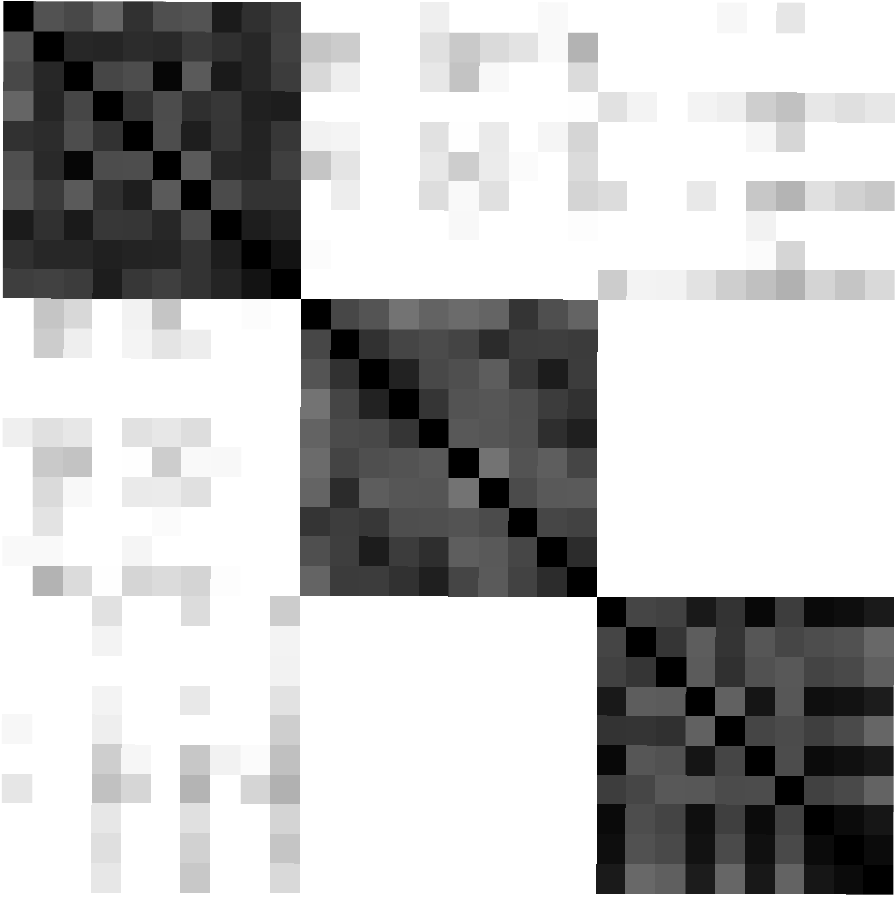}
        \end{minipage}}
        \hspace{-11.4mm}
          \subfigure[NODL]{
        \begin{minipage}[a]{0.24\textwidth}
          \centering
    \includegraphics[width =95pt ,height =85pt]{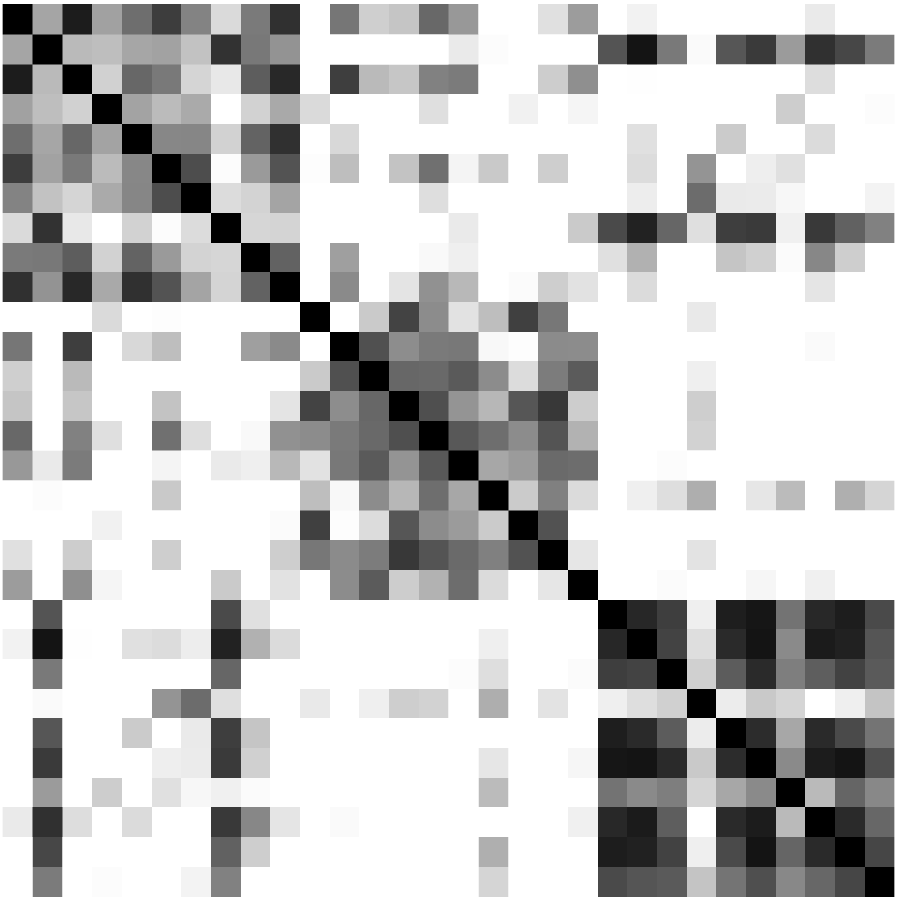}
        \end{minipage}}
        \hspace{-11.4mm}
        \subfigure[ELLA-Rand]{
        \begin{minipage}[a]{0.24\textwidth}
          \centering
    \includegraphics[width =95pt ,height =85pt]{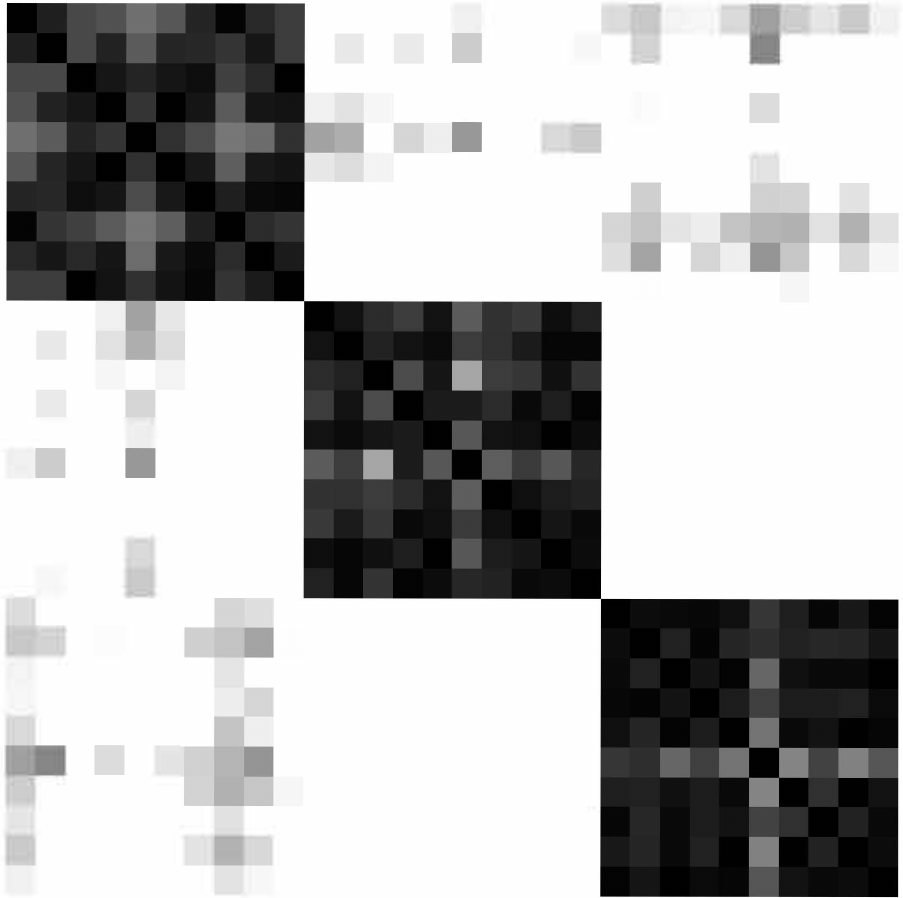}
        \end{minipage}}
        \hspace{-11.4mm}
       \subfigure[ELLA-Info]{
        \begin{minipage}[a]{0.24\textwidth}
          \centering
    \includegraphics[width =95pt ,height =85pt]{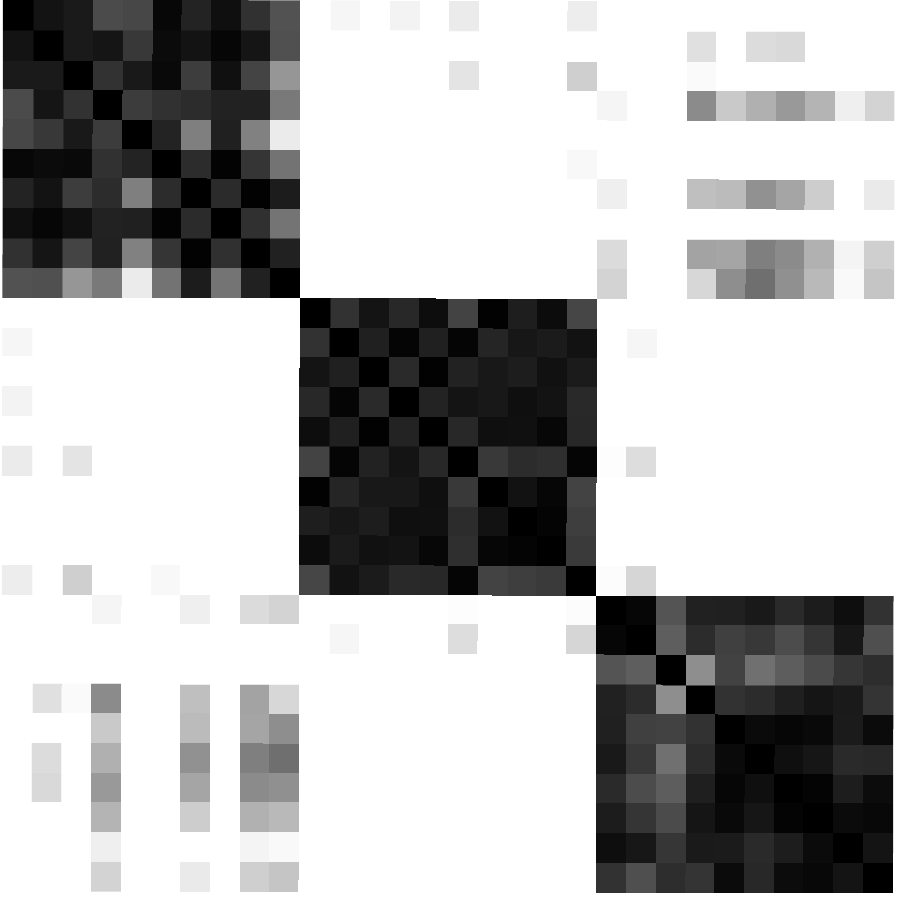}
        \end{minipage}}
        \hspace{-11.4mm}
       \subfigure[Our $\mr{FCL^3}$-Rand]{
        \begin{minipage}[a]{0.24\textwidth}
          \centering
    \includegraphics[width =95pt ,height =85pt]{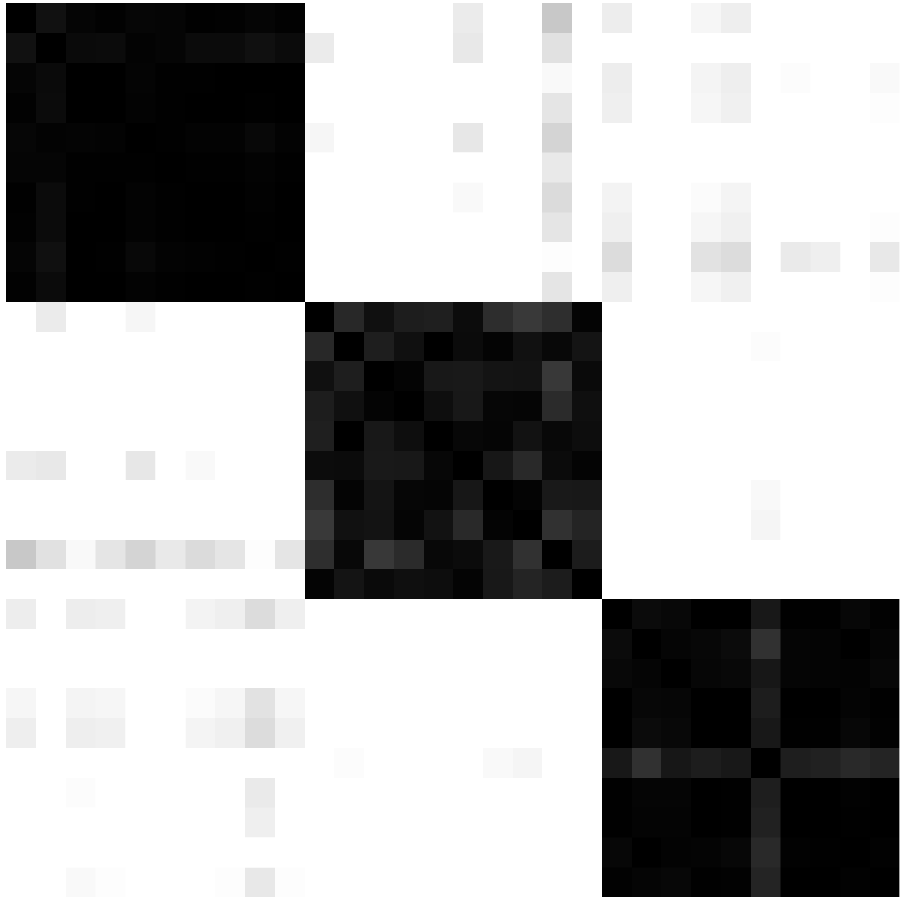}
        \end{minipage}}
        \hspace{-8.4mm}
   \caption{The correlation matrices of \textbf{Disjoint} dataset for different models: (a) Ground Truth, (b) STL, (c) CMTL, (d) DG$\_$MTL, (e) AMTL, (f) FCMTL, (g) NODL, (h) ELLA-Rand, (i) ELLA-Info and (j) our proposed $\mr{FCL^3}$ models. The darker color indicates the higher correlation. There are 30 tasks clustered into 3 clusters in the ground truth, where each task has 40 dimensions, and 25-25 training-test samples are used in each task.}
   \vspace{-10pt}
  \label{fig:disjointdata}
\end{figure*}

\noindent\textbf{Parkinson} dataset\footnote{https://archive.ics.uci.edu/ml/datasets/parkinsons+telemonitoring}: this dataset consists of Parkinson's disease symptom score of 5,875 observations for 42 patients. Each task is a symptom score prediction problem of a patient, and each sample consists of 16 biomedical features. Then total task number is 42, and the number of samples for each task (patient) varying from 101 to 168. Even though the response of this dataset has two scores, i.e., Total and Motor, we establish one regression dataset in this experiment: Parkinson-Motor, where Parkinson-Total has similar performance trend as Parkinson-Motor.

\noindent\textbf{Landmine} dataset: this dataset is used to detect whether a land mine is presented in an area based on radar images or not. It can thus be modeled as a binary classification problem. Each object in this dataset is described using a 9-dimensional feature vector (i.e., three correlation-based features, four-moment based features, one spatial variance feature and one energy-ratio feature), and its corresponding binary label (1 for landmine and -1 for clutter). The task number is 29 after dividing the total of 14,820 samples into 29 different geographical regions.

\noindent\textbf{SmartMeter} dataset \footnote{http://www.ucd.ie/issda/data/commissionforenergyregulationcer/}: this dataset is collected by the Irish CER during a smart metering trial conducted in Ireland, and the target is to research how the consumption impact on the household characteristics. In this experiment, we adopt the provided 81-length feature vectors of electricity consumption data (such as daily consumption figures, statistical aspects, etc) for each household \cite{sun2017joint,cong2018user}, and the number of characteristics (such as cooking style, household income, etc) from questionnaires is 16. We model each characteristic as a separate task, and the task number is 16.

\noindent\textbf{Caltech-Birds} dataset\footnote{http://www.vision.caltech.edu/visipedia/CUB-200-2011.html}: this image dataset containing 200 categories is a fine grained bird classification problem. We thus treat this dataset as a multi-task learning problem, and each task can be a classification task with one class against some negative samples. More specifically, we run several comparisons among 5 categories (i.e., Grebe, Cormorant, Blackbird, Kingfisher and Gull), which are composed of 24 bird subcategories in total. The example images are shown in Fig.~\ref{fig:example_image}. To better represent each image, a 128-dimensional deep feature for each image is extracted using the VGG model\cite{chatfield2014return}.

\begin{figure}[t]
  \subfigure{
        \begin{minipage}[a]{0.24\textwidth}
          \centering
    \includegraphics[trim = 0mm 0mm 0mm 0mm, clip, width =245pt, height=110pt]{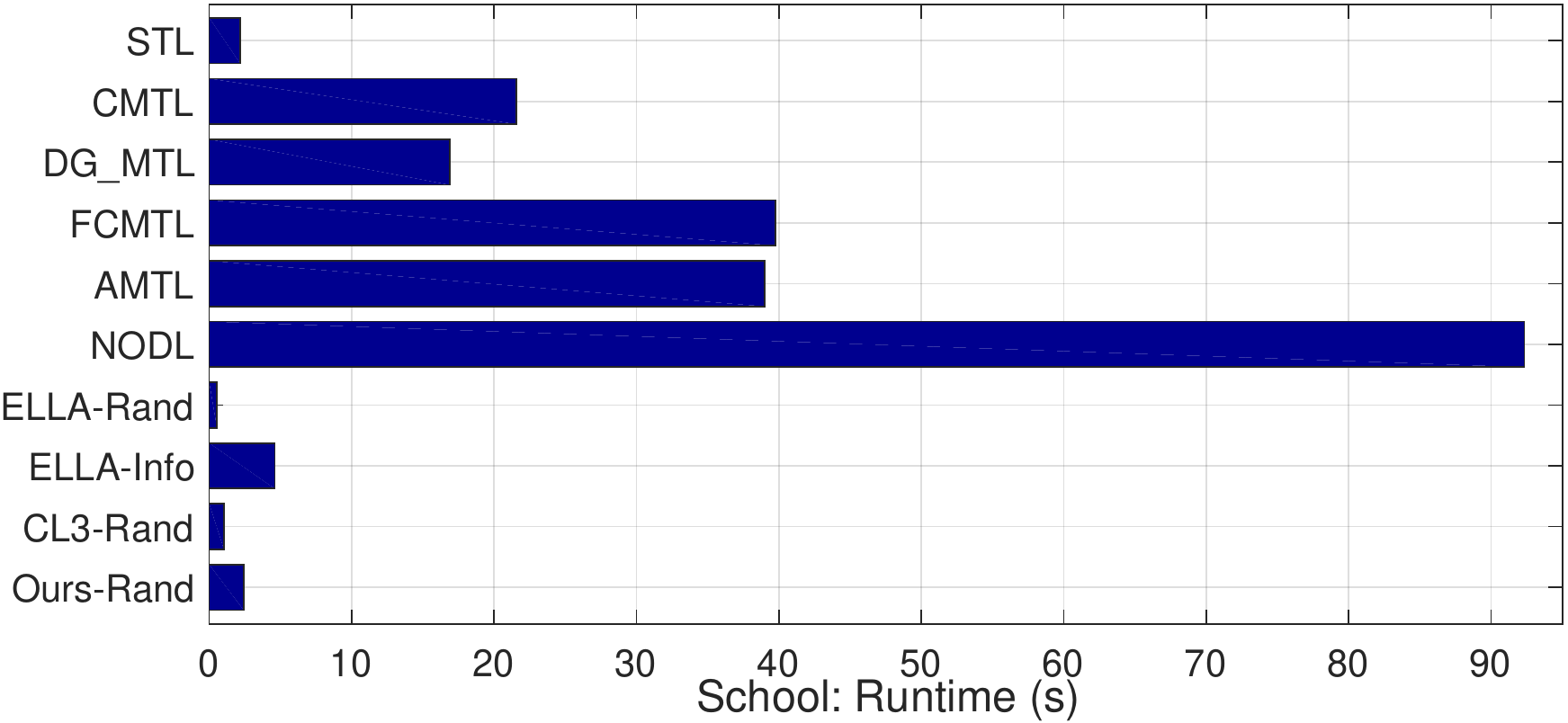}
        \end{minipage}}
        \vspace{-3.5mm}\\
    \subfigure{
           \begin{minipage}[a]{0.24\textwidth}
          \centering
    \includegraphics[trim = 0mm 0mm 0mm 0mm, clip, width =245pt, height=115pt]{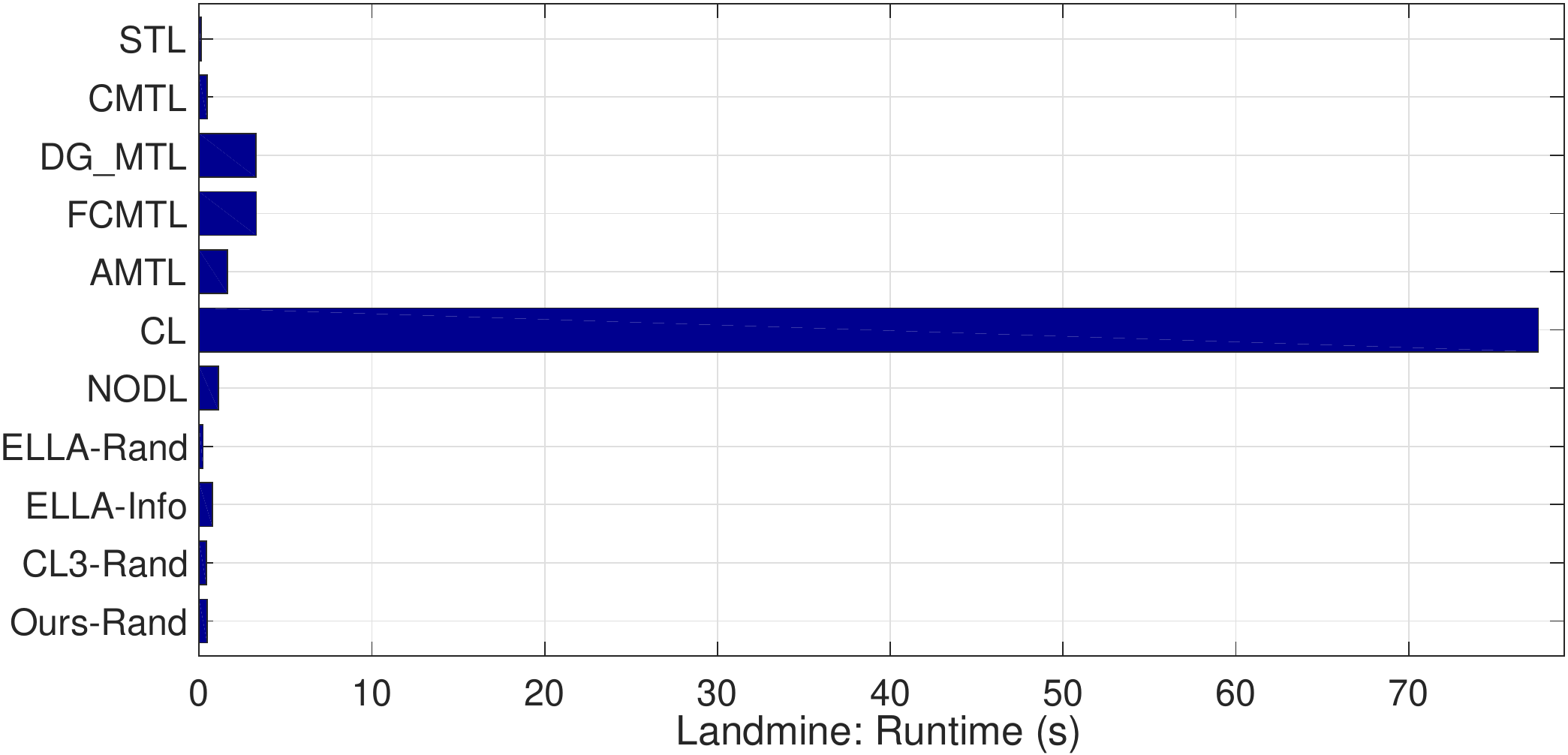}
         \end{minipage}}
         \hspace{-3.4mm}
         \vspace{-4.0mm}
         \caption{The demonstration of computational time (seconds) on School (\textbf{Top}) and Landmine (\textbf{Bottom}) datasets, where each bar denotes each competing models.}
        \label{fig:runtime} 
\end{figure}




All the used datasets in this experiment are normalized, and more details are provided in Table~\ref{table:dataset}. In our experiments, we randomly divide each task into 50$\%$-50$\%$ training-test set, and the experimental results averaged over ten random repetitions are presented in Table~\ref{table:regression&classification_result}, Table~\ref{table:classification_accuracy_result}, Fig.~\ref{fig:disjointdata} and Fig.~\ref{fig:runtime}. Based on the presented results, several observations are as follows:

\textbf{1)} For the regression problems, our $\mr{FCL^3}$ model performs better than the state-of-the-art MTL methods and achieves $0.047$, $0.035$ and $0.164$ improvement in terms of RMSE on the Disjoint, Parkinson-Motor and School datasets, respectively. This indicates that better performance can be obtained by simultaneously capturing task clustering structure and leaning more discriminative features. Comparing with lifelong learning models, e.g., NODL and ELLA-Rand, it can be seen from the Table~\ref{table:regression&classification_result} that our flexible clustered lifelong learning model could benefit from the underlying task cluster structure. Meanwhile, our $\mr{FCL^3}$ model outperforms our previous conference work due to the self-reconstruction constraint of the autoencoder architecture. Another interesting thing is that NODL (learning dictionary incrementally) is clearly worse than ELLA-Rand. One reason is that although it is an online dictionary learning which can adapt the dictionary structure via continuous birth and death, it 1) ignores the previously learned sparse coefficients and 2) involves more redundancy information when the number of dictionary element becomes large. Moreover, all multi-task learning (MTL) and lifelong learning models outperform single task learning on these three regression datasets. In addition, for the Disjoint dataset, the correlation matrices of the obtained model parameters are provided in Fig.~\ref{fig:disjointdata}. Notice that our proposed $\mr{FCL^3}$ model can capture the tasks cluster structure well when comparing with other MTL models, such as CMTL.

\textbf{2)} For the classification problems, our $\mr{FCL^3}$ model obtains corresponding $0.512\%$ and $1.597\%$ improvement in terms of AUC on the SmartMeter and Landmine datasets. However, both CMTL and FCMTL achieves slightly better performance than our $\mr{FCL^3}$ model on Caltech-Bird dataset. The possible reasons are that 1) deep feature extracted by VGG model for Caltech-Bird dataset is more discriminative, and several rich features may be damaged after mapped by the feature learning library; 2) both CMTL and FCMTL can jointly learn all the tasks in the offline regime, whereas our $\mr{FCL^3}$ model can just learn the encountered tasks gradually. Furthermore, both ELLA-Rand and ELLA-Info perform a little worse than our $\mr{FCL^3}$ model on most datasets, this similar observation is because that ELLA model treats the observed tasks independently and further neglects the underlying clustering structural information, whereas our $\mr{FCL^3}$ model can establish a clustered lifelong learning system by incorporating flexible task clustering structure. Additionally, we have the similar observation as the regression problems when comparing with our conference work.

As the accuracy result shown in Table.~\ref{table:classification_accuracy_result}, our $\mr{FCL^3}$ model can also achieve corresponding $1.009\%$ improvement in terms of ACC on the SmartMeter dataset. Comparing with several clustered multi-task learning and lifelong learning models, we have find that our $\mr{FCL^3}$ model outperforms others significantly. The reason why we do not test the Landmine and Caltech-Birds datasets in this experiment is that both our used Landmine and Caltech-Birds datasets have highly biased class distributions. The classification accuracy would only be informative for specific applications with well-specified tradeoffs between true and false positives.

\begin{table}[t]
\caption{Comparisons between our $\mr{FCL^3}$ model and state-of-the-arts in terms of accuracy on SmartMeter dataset: mean and standard errors averaged over ten random runs. Models with the best performance are bolded.}
\vspace{-5.0pt}
\centering
\scalebox{0.83}{
\begin{tabular}{|c|c|c|c|c|c|c|c|}
\hline
 {Models}&Evaluation&  SmartMeter Dataset  \\
 \hline
 STL &Acc($\%$)& 66.858$\pm$0.30  \\ \hline

 CMTL\cite{Zhou:2011}& Acc($\%$)&70.785$\pm$0.12  \\ \hline

 DG$\_$MTL\cite{Kang:2011} &Acc($\%$)&69.723$\pm$0.53   \\ \hline

 FCMTL\cite{zhou2016flexible} &Acc($\%$)&   70.752$\pm$0.12     \\ \hline

 AMTL \cite{lee2016asymmetric} &Acc($\%$)&67.829$\pm$0.15  \\ \hline

 CL\cite{pentina2015curriculum} &Acc($\%$) & 69.094$\pm$0.53  \\ \hline
NODL\cite{Garg2017Neurogenesis} &Acc($\%$) &67.784$\pm$1.02 \\ \hline
 ELLA-Rand\cite{ruvolo2013ella} &Acc($\%$) &   70.184$\pm$0.11  \\  \hline

 ELLA-Info\cite{ruvolo2013active}  &Acc($\%$) &  70.703$\pm$0.24 \\ \hline

 CL3-Rand\cite{gan2018clusteredlifelong}&Acc($\%$)&  70.998$\pm$0.13  \\ \hline

 Ours-Rand  &Acc($\%$)&\textbf{71.761$\pm$0.15}\\ \hline

 \end{tabular}
}
\label{table:classification_accuracy_result}
\end{table}

\begin{figure}[t]
\center
  \hspace{-40.6mm}
    \subfigure{
        \begin{minipage}[a]{0.24\textwidth}
          \centering
    \includegraphics[width =232pt,height =145pt]{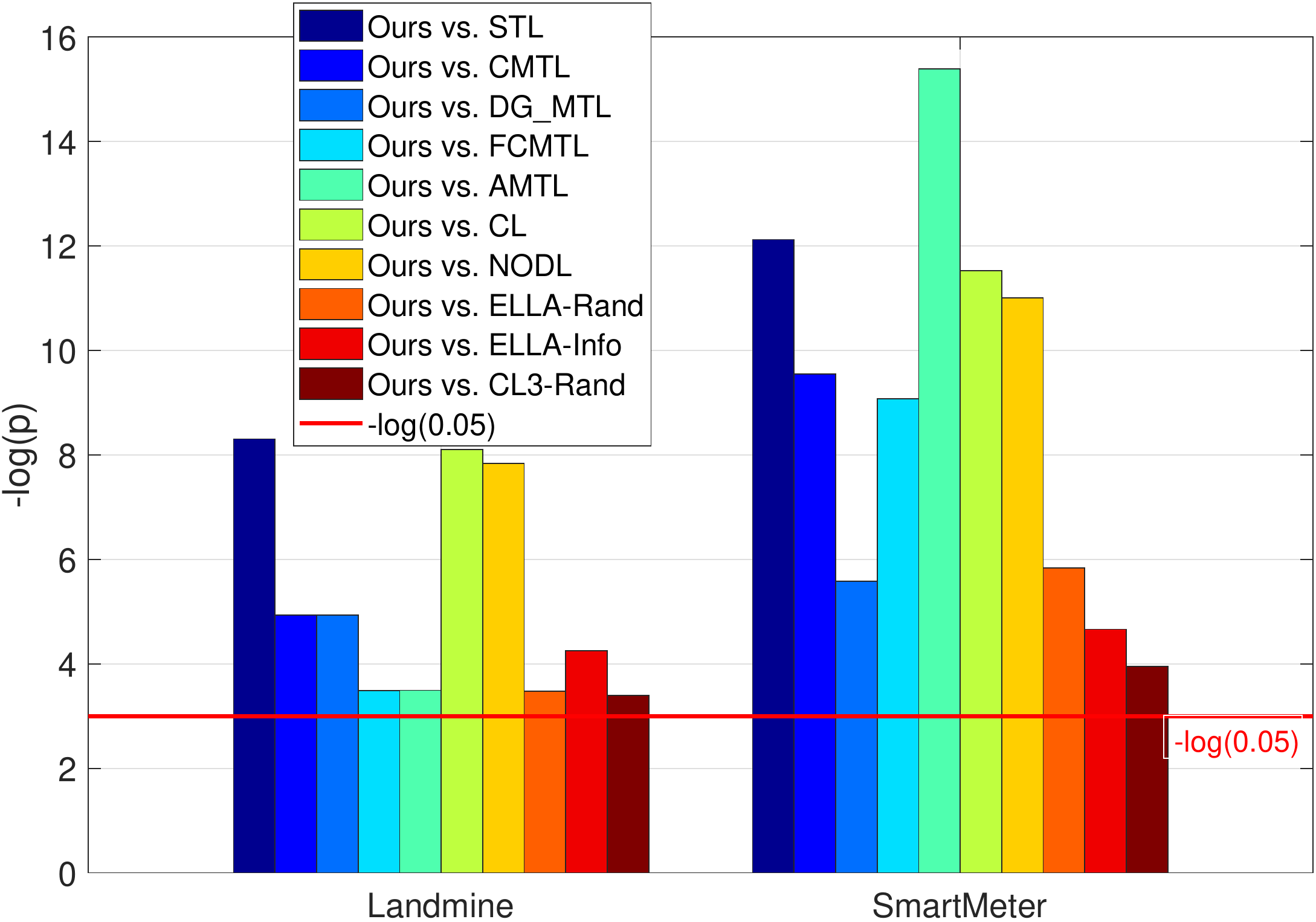}
         \end{minipage}}
         \vspace{-10pt}
    \caption{p-value of t-test between our model and others on both Landmine and SmartMeter datasets. We pre-process using $-\mr{log}(p)$ so that the large value shown in this figure denotes the more significance of our model compared with others.}
    \label{fig:ttest}
\end{figure}

In order to demonstrate the statistical significance of our proposed model, we present a significance test (t-test) for the classification results shown in Table~\ref{table:regression&classification_result}. Specifically, we adopt a significance level of 0.05, i.e., when the $p$-value is less than 0.05, the performance difference of two models is statistically significant. As the classification results shown in Fig.~\ref{fig:ttest}, we use the $-\mr{log}(p)$ processing on $p$-value, and the comparison shows that our framework outperforms others significantly if the values are greater than $-\mr{log}(0.05)$. Obviously, our proposed model performs significantly on both Landmine and SmartMeter datasets. Since the performance of our model is not the best on the Caltech-Birds, the statistical significance of this dataset is not presented in Fig.~\ref{fig:ttest}.

\textbf{3)} For the comparison in terms of time consumption, we present the corresponding runtimes of both School and Landmine datasets in Fig.~\ref{fig:runtime}. As shown in Fig.~\ref{fig:runtime}, our $\mr{FCL^3}$ model is more efficient than several competing MTL models, e.g., CMTL and AMTL. However, our $\mr{FCL^3}$ model is little slower than lifelong learning model: ELLA-Rand. The reason is that our $\mr{FCL^3}$ model needs to update the model knowledge library, whereas ELLA-Rand donot need. The reason why our model is a little slower than CL3-Rand is that our $\mr{FCL^3}$ model adopts an autoencoder architecture to learn the feature. In addition, the computational time for CL~\cite{pentina2015curriculum} is very higher, it is because that CL model has to find the best learning order on the remaining tasks, which needs more runtime. All the experiments are implemented using Matlab on computer with 8G RAM, Intel i7 CPU.

\begin{figure}[t]
\centering
\includegraphics[width =120pt ,height =96pt]{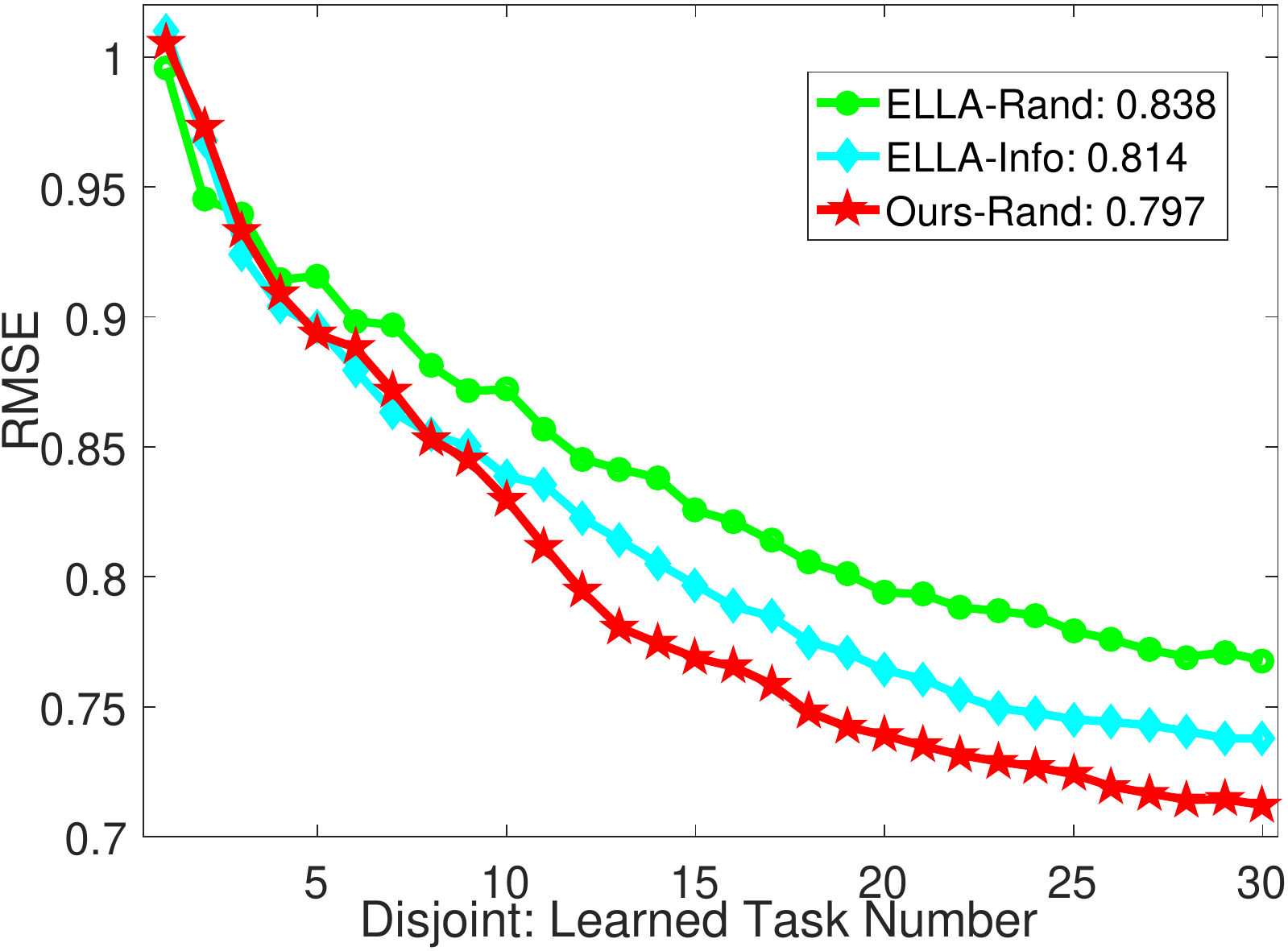}
\includegraphics[width =120pt ,height =96pt]{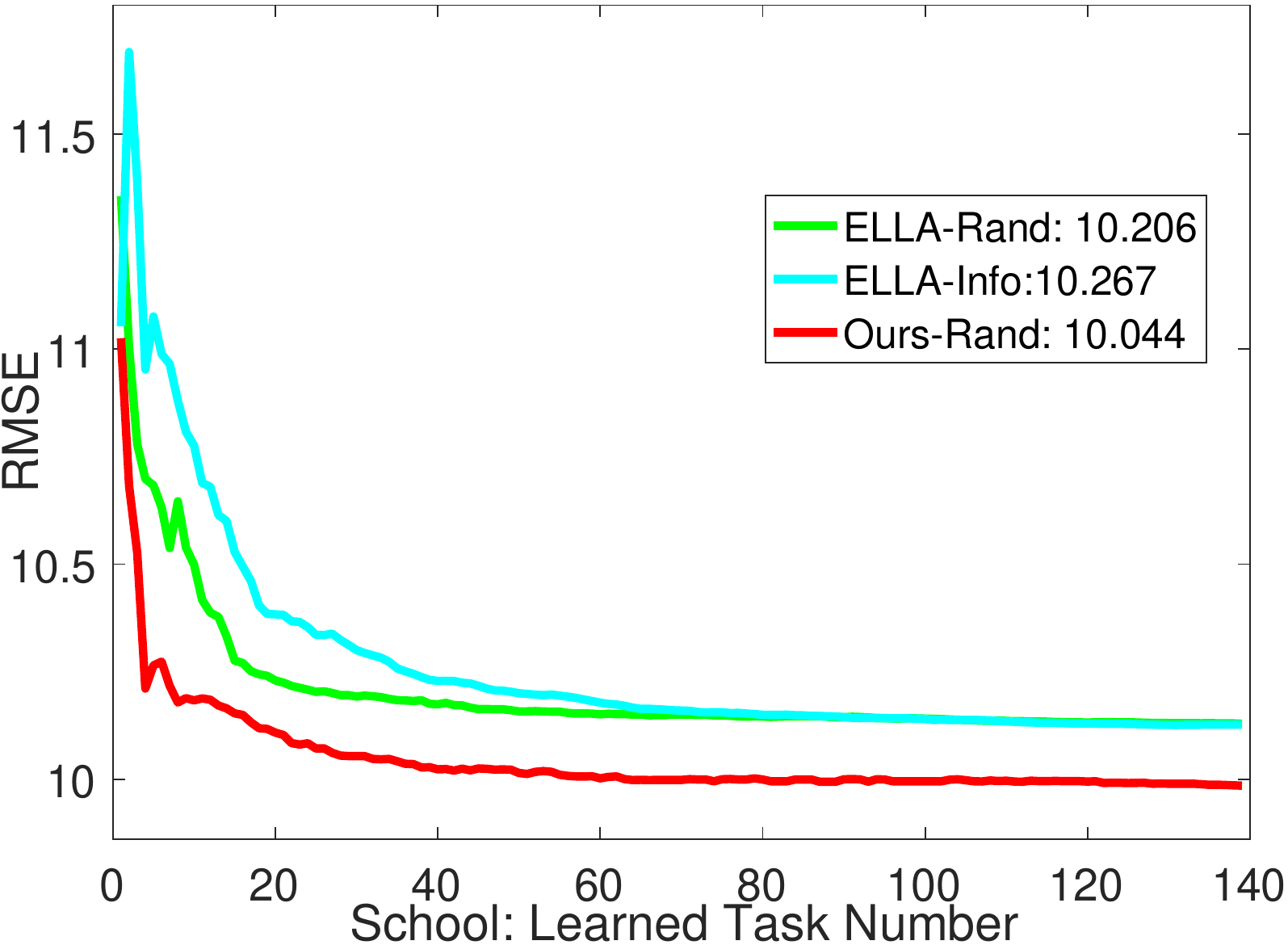}
\vspace{-10.0pt}
\caption{Regression comparisons with ELLA-Rand, ELLA-Info and Ours-Rand on Disjoint and School datasets. Along horizontal and vertical axes are the number of learned task and value of RMSE. The corresponding legend in each figure provides the mean performance of each curve (model).}
\vspace{-10.0pt}
\label{fig:regression_learnedtask}
\end{figure}


\begin{figure}[h]
\centering
\includegraphics[width =120pt ,height =96pt]{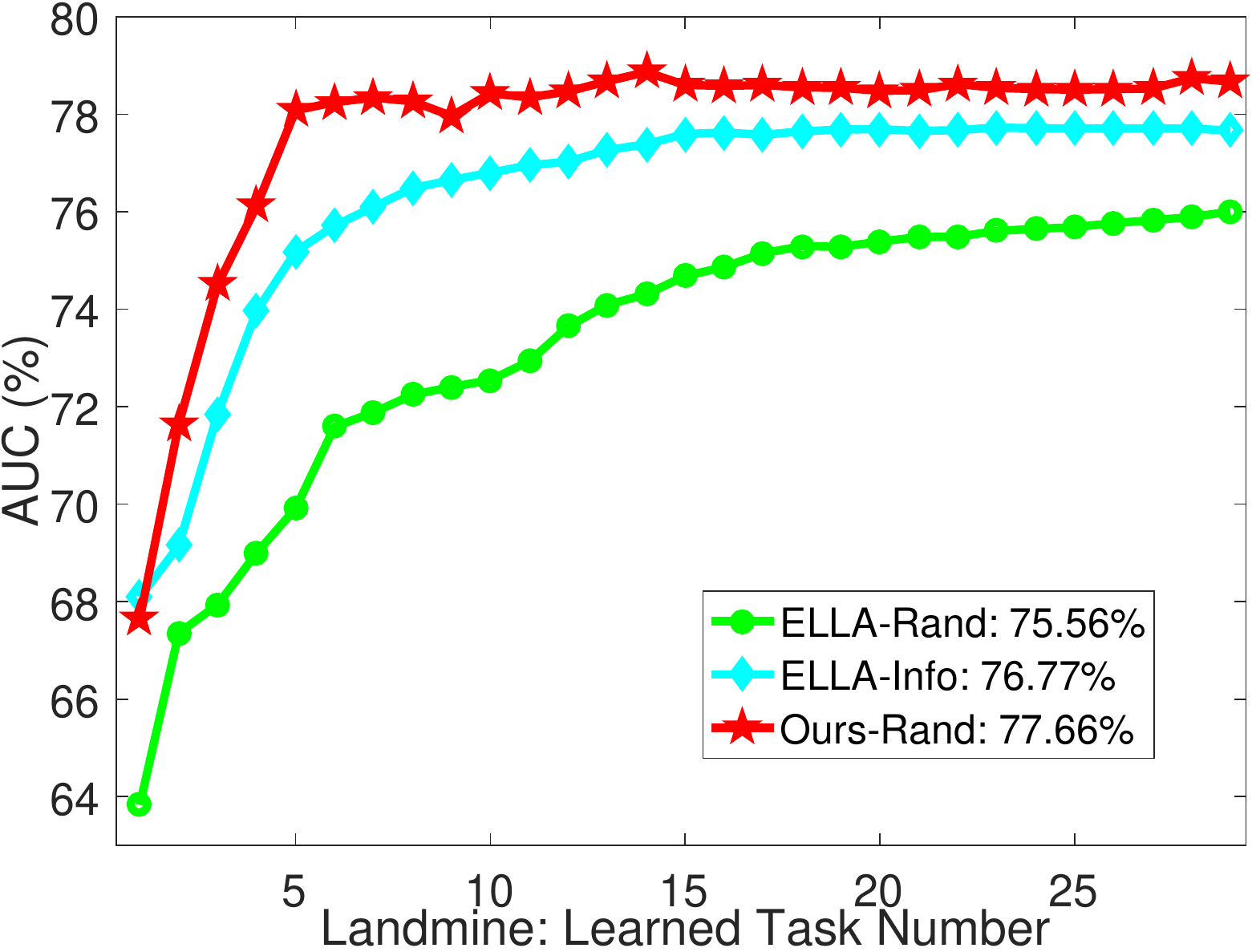}
\includegraphics[width =120pt ,height =96pt]{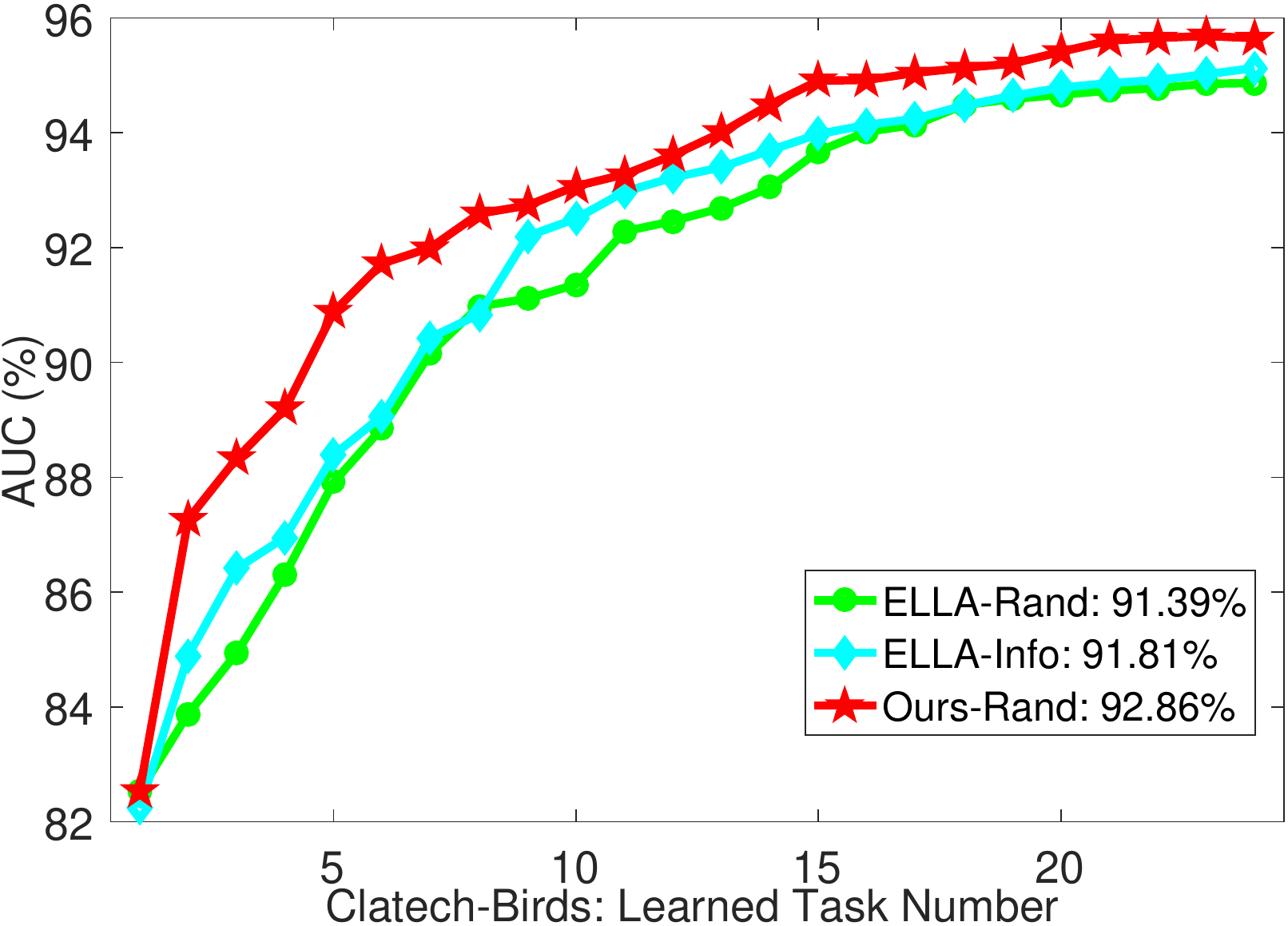}
\vspace{-10.0pt}
\caption{Classification comparisons with ELLA-Rand, ELLA-Info and Ours-Rand on Landmine and Caltech-Bird datasets. Along horizontal and vertical axes are the number of learned task and value of AUC ($\%$). The corresponding legend in each figure provides the mean performance of each curve (model).}
\vspace{-10.0pt}
\label{fig:classification_learnedtask}
\end{figure}

\subsection{Comparison on the Number of Learned Tasks}
In addition to the experimental results in the Table \ref{table:regression&classification_result}, in this subsection, we explore how the number of learned tasks influence the regression and classification performance of our $\mr{FCL^3}$ model. More specifically, School, Disjoint, Landmine and Caltech-Birds datasets are used by randomly dividing into $50\%-50\%$ training-test set, and each learned task can be reconstructed via $Ds_i$, where $i\in \{1,\ldots,t\}$. From the performance curves presented in Fig.~\ref{fig:regression_learnedtask} and Fig.~\ref{fig:classification_learnedtask}, we can notice that the performance (averaged over several runs) of our $\mr{FCL^3}$ model can tend to be better than ELLA-Rand and ELLA-Info when new tasks arrives at the lifelong learning system, the main reason is that: 1) we adopt an additional model knowledge library $S$ in the $\mr{FCL^3}$ model, i.e., our knowledge library (i.e., feature learning library $\{D, L\}$ and model knowledge library $S$ could gradually accumulate more useful knowledge than ELLA-Rand; 2) when strange task environments come, our proposed $\mr{FCL^3}$ model can judge new task environment and further added it into model knowledge library progressively, i.e., more structural information from real-world data can be captured.

\begin{figure}[t]
\centering
\includegraphics[width =235pt ,height =91pt]{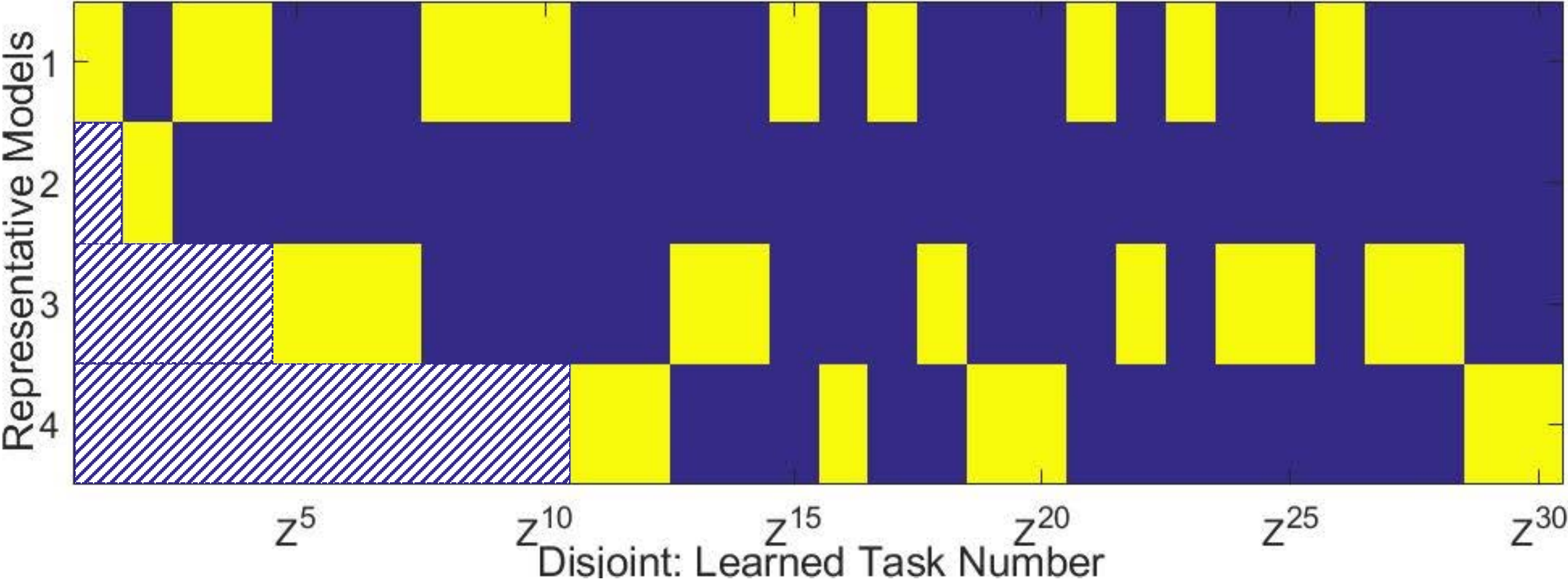}\\
\vspace{3.pt}
\includegraphics[width =235pt ,height =91pt]{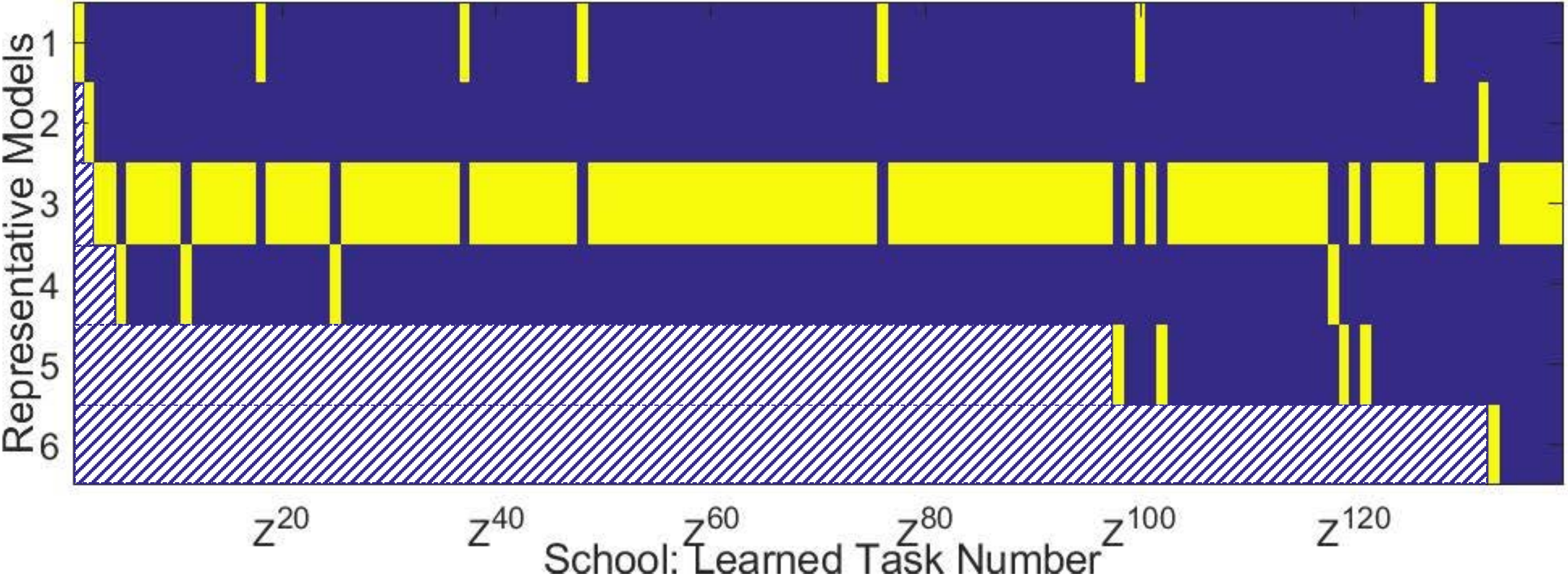}\\
\vspace{3.pt}
\includegraphics[width =235pt ,height =91pt]{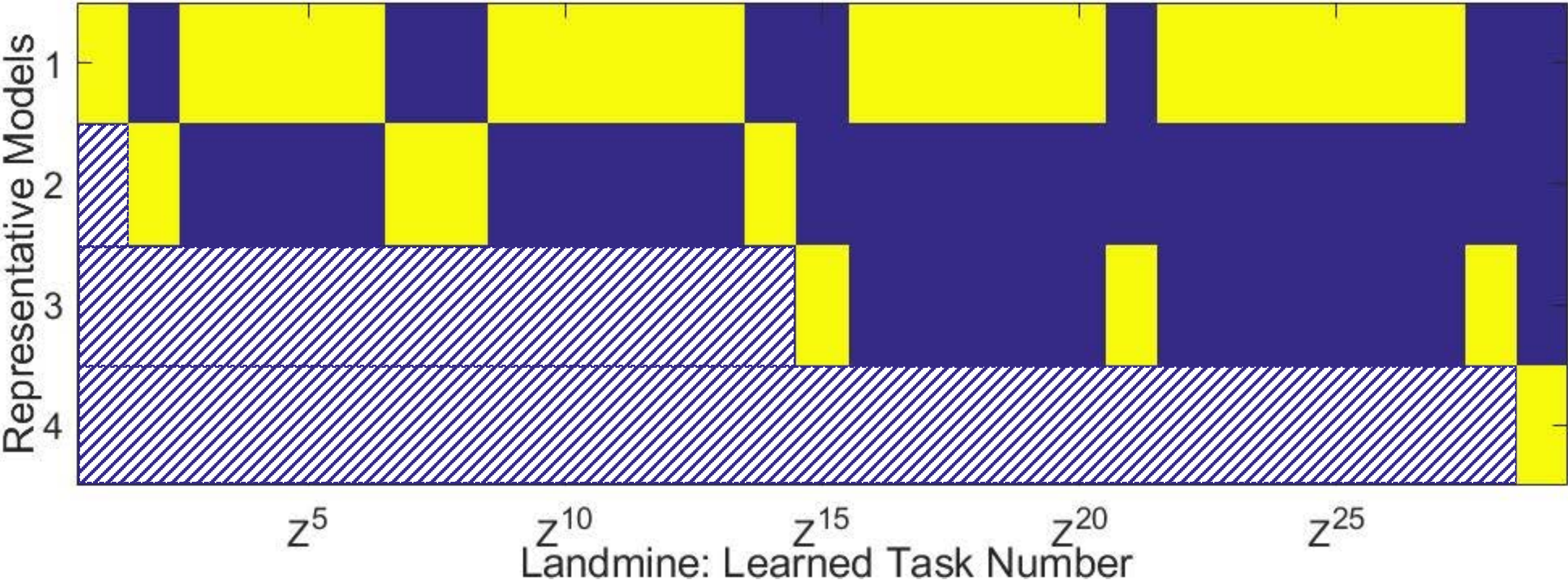}\\
\vspace{3.pt}

\includegraphics[width =235pt ,height =91pt]{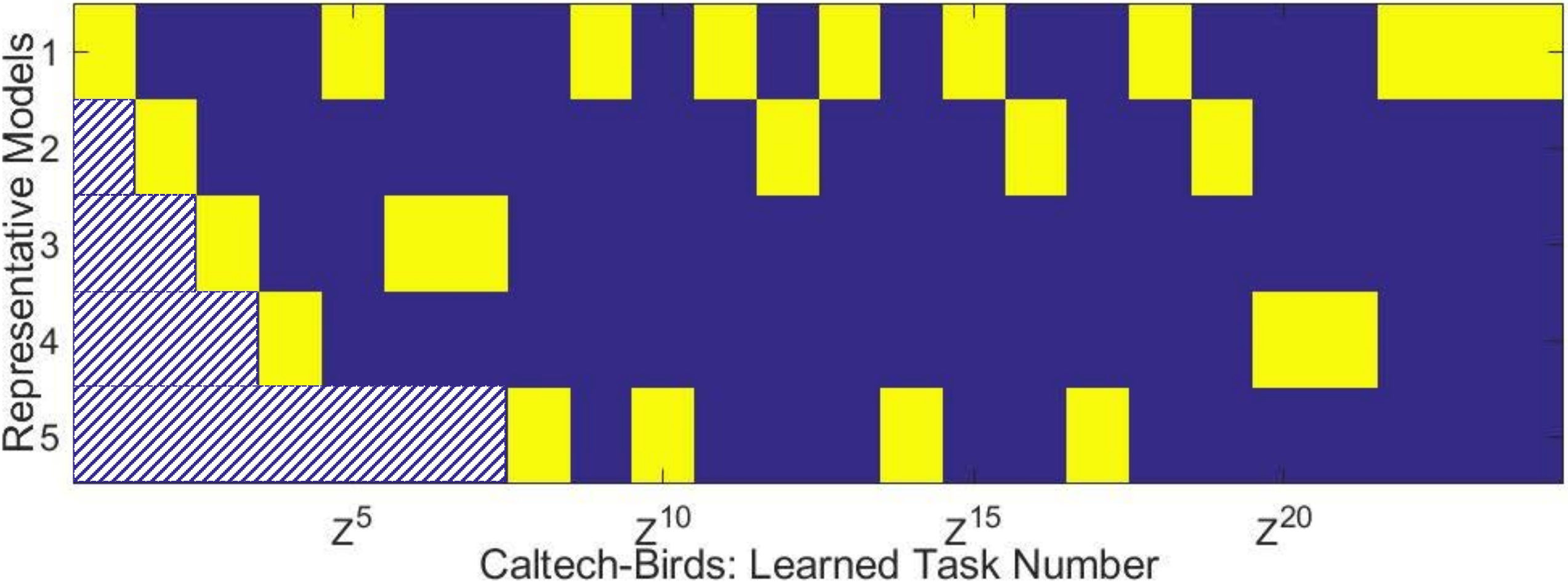}
\vspace{-8.0pt}
\caption{Demonstration of the corresponding vector ${Z}^t$'s and representative models obtained by our proposed $\mr{FCL^3}$ model (from top to bottom: Disjoint, School, Landmine and Caltech-Birds datasets), where masked regions denote NaN, and yellow block and blue block are corresponding max probability value or not, i.e., representative models or not.}
\vspace{-10.0pt}
\label{fig:representatives}
\end{figure}


\subsection{New Task Environments in Real-world Datasets}
In this subsection, we investigate how many task environments (clusters) in our adopted real-world datasets. Specifically, we plot the corresponding values in $Z^t$ (i.e., Eq.~\eqref{eq:subproblem_z}) for each new task among School, Disjoint, Caltech-Birds and Landmine datasets. Meanwhile, we run this experiment by randomly dividing the dataset into $50\%-50\%$ training-test sets, and the obtained results are given in Fig.~\ref{fig:representatives}, where the max probability value in $Z^t$'s are marked as yellow block. As shown in  Fig.~\ref{fig:representatives},  the first yellow block emerging in each new row indicates that the max probability value in $Z^t$ is $z_{K_t+1}$ (i.e., $e$ in Eq.~\eqref{eq:representive_task1}) for this new task, i.e., a new task environment (i.e., representative model) appears. We can notice that the new task environments for each dataset could be self-selected step-by-step. For example, for the Caltech-Bird dataset, the representative model are marked as the $1$-th, $2$-th, $3$-th, $4$-th and $8$-th coming tasks. The corresponding task environment numbers for Disjoint, School, Landmine and Caltech-Bird datasets are 4, 6, 4 and 5. Intuitively, the task environment number for Caltech-Birds dataset is $5$, which is in accordance with the category task number we used in this paper; even though the task environment number for Disjoint dataset we find is $4$, the outlier task number should be 1 (i.e., one yellow block appears in the second row for Disjoint dataset). This observation can also support the effectiveness of our $\mr{FCL^3}$ model.



\begin{table}[htbp]
\caption{Comparisons of the effect of task order on Disjoint and Landmine datasets.}
\vspace{-7.0pt}
\centering
\scalebox{0.86}{
\begin{tabular}{|c|c|c|c|c|c|}
\hline
 {}& Evaluation&ELLA-Rand\cite{ruvolo2013ella}&Ours-OneByOne & Ours-Rand  \\
 \hline\hline
Disjoint        & RMSE&0.767$\pm$0.03     &0.747$\pm$0.02  &0.713$\pm$0.01       \\ \hline
Landmine     &AUC($\%$) & 77.429$\pm$0.79   &78.284$\pm $0.12  &79.022$\pm$0.42      \\   \hline
 \end{tabular}
}
\vspace{-10.0pt}
\label{table:taskorder}
\end{table}

\subsection{Effect of the Task Order}
To study how the task order affect the generalization performance of our proposed $\mr{FCL^3}$ model, we adopt Disjoint and Landmine datasets by randomly dividing these data into $50\%-50\%$ training-test set for this experiment, and record the experiment results in Table.~\ref{table:taskorder}. From the presented results, we can observe that the performance of Ours-OneByOne is similar to that of Ours-Rand on the Landmine dataset, but isnot on the Disjoint dataset (the performance of Ours-Rand is better than that in Ours-OneByOne). It is because that the representative models of Landmine dataset tend to be random distribution, while the representative models of Disjoint are following the uniform distribution (i.e., cluster center is in a one-by-one way, where each cluster includes 10 regression tasks). This observation indicates that our proposed $\mr{FCL^3}$ model can be applied into the lifelong learning system, which has not enough prior knowledge about the task order or their distribution. Additionally, the performance of Ours-OneByOne is better than ELLA-Rand on these two datasets, which also verifies the effectiveness of the representative models, i.e., model knowledge library.

\begin{figure}[htbp]
    \subfigure[Disjoint Dataset]{
        \begin{minipage}[a]{0.24\textwidth}
          \centering
    \includegraphics[trim = 0mm 0mm 0mm 0mm, clip, width =125pt ,height =108pt]{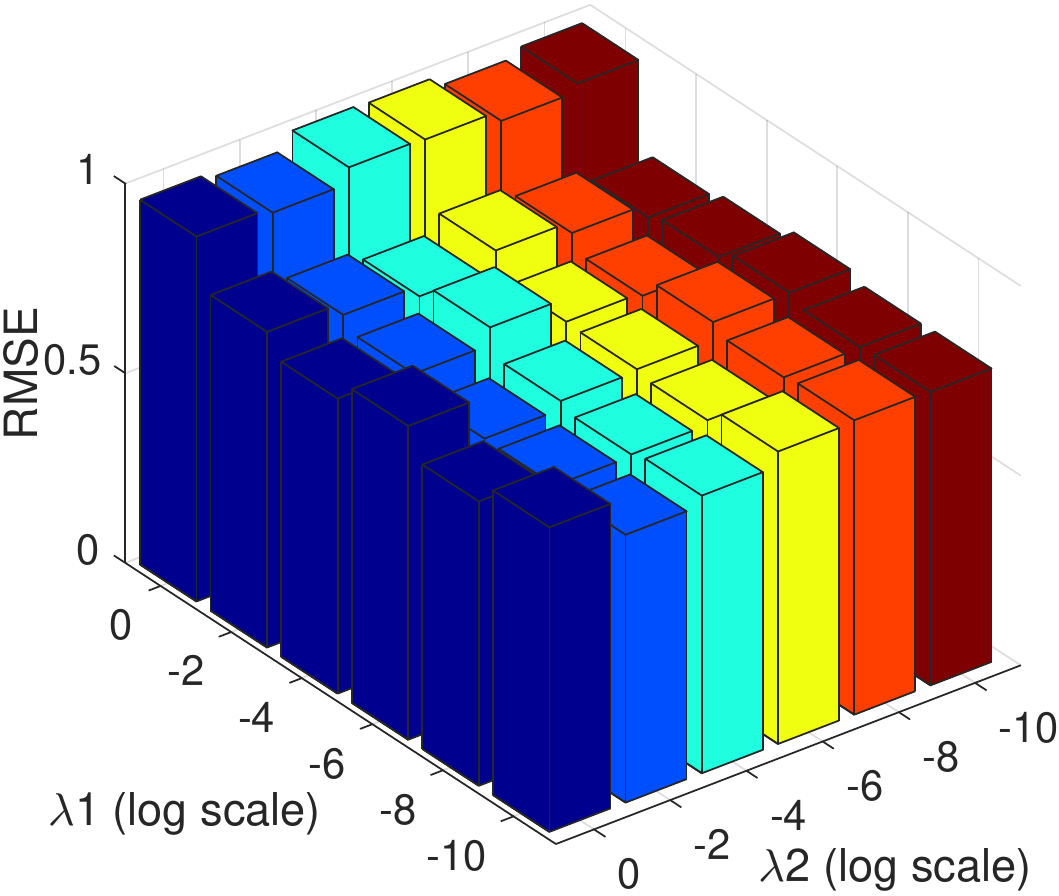}
         \end{minipage}}
         \hspace{-2.4mm}
    \subfigure[School Dataset]{
        \begin{minipage}[a]{0.24\textwidth}
          \centering
    \includegraphics[trim = 0mm 0mm 0mm 0mm, clip, width =125pt ,height =108pt]{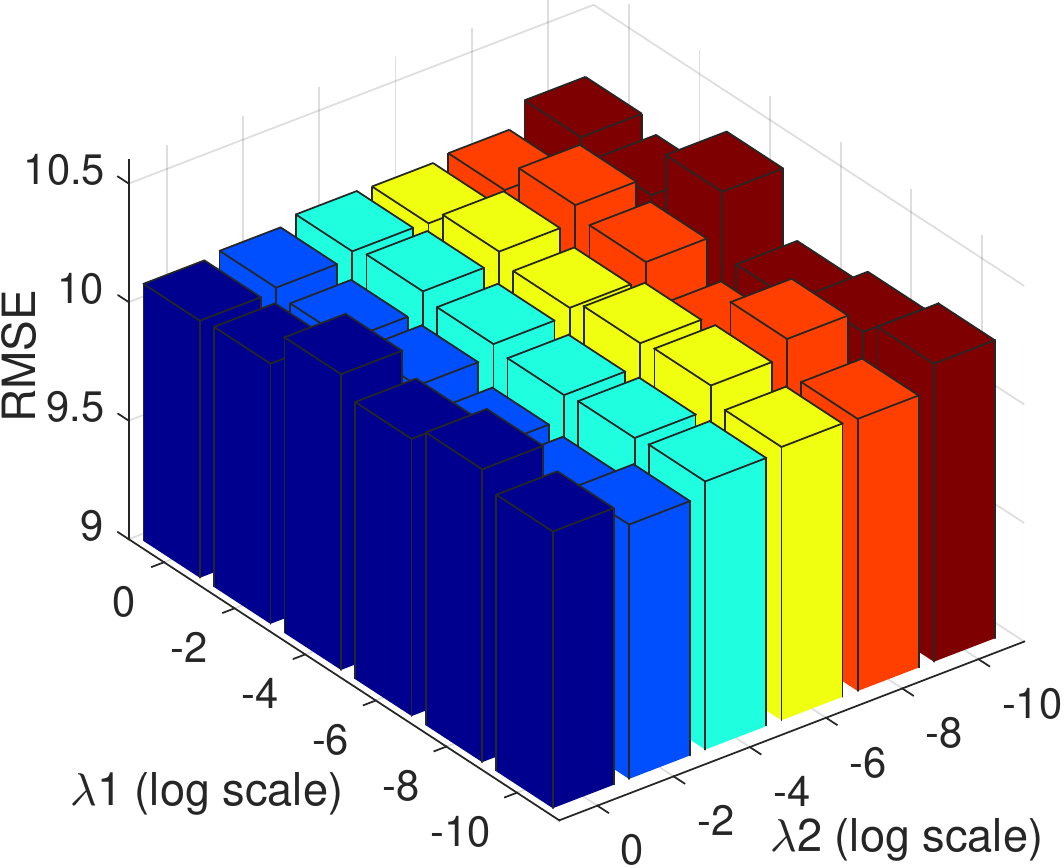}
        \end{minipage}}
        \vspace{-6.0mm}
         \caption{The effect of the regularization parameters $\lambda_1$ and $\lambda_2$ on Disjoint (a) and School (b) datasets, respectively.}
         \vspace{-10.0pt}
        \label{fig:Effect_Parameter_Regression} 
\end{figure}

\begin{figure}[htbp]
         \subfigure[Landmine Dataset]{
        \begin{minipage}[a]{0.24\textwidth}
          \centering
    \includegraphics[trim = 0mm 0mm 0mm 0mm, clip, width =125pt ,height =108pt]{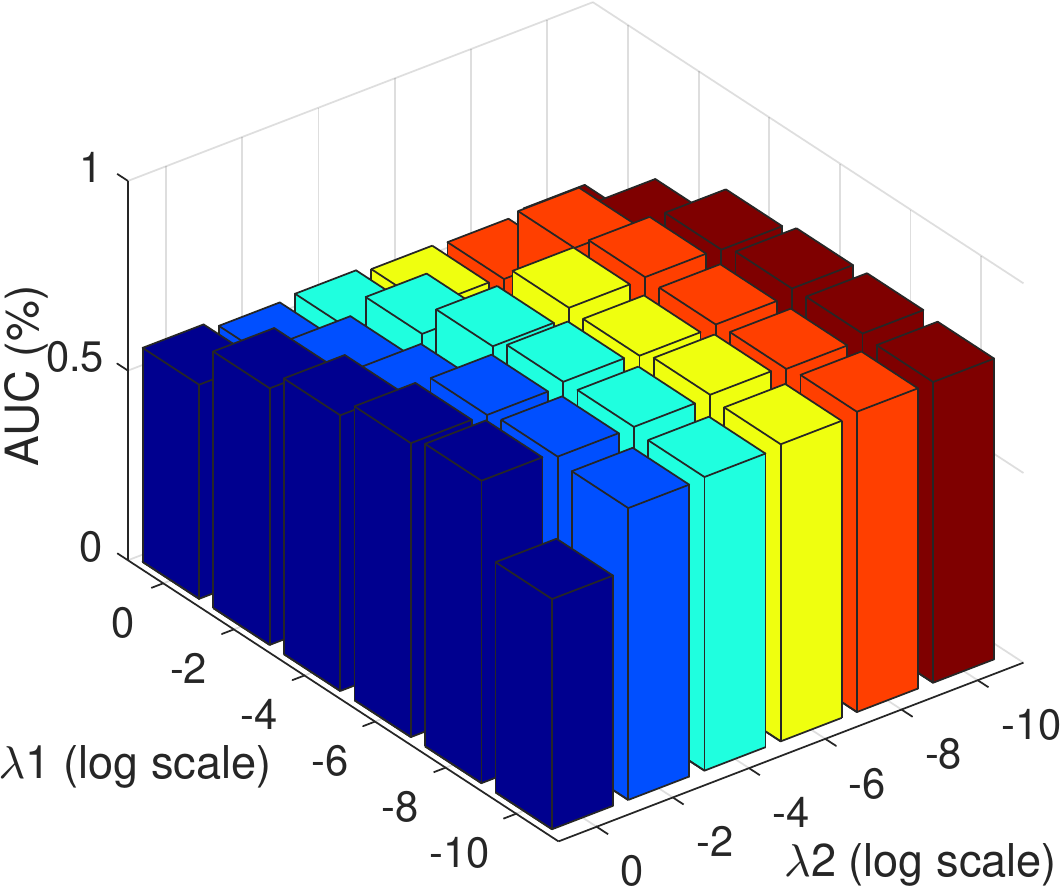}
         \end{minipage}}
         \hspace{-2.4mm}
    \subfigure[Caltech-Birds Dataset]{
        \begin{minipage}[a]{0.24\textwidth}
          \centering
    \includegraphics[trim = 0mm 0mm 0mm 0mm, clip, width =125pt ,height =108pt]{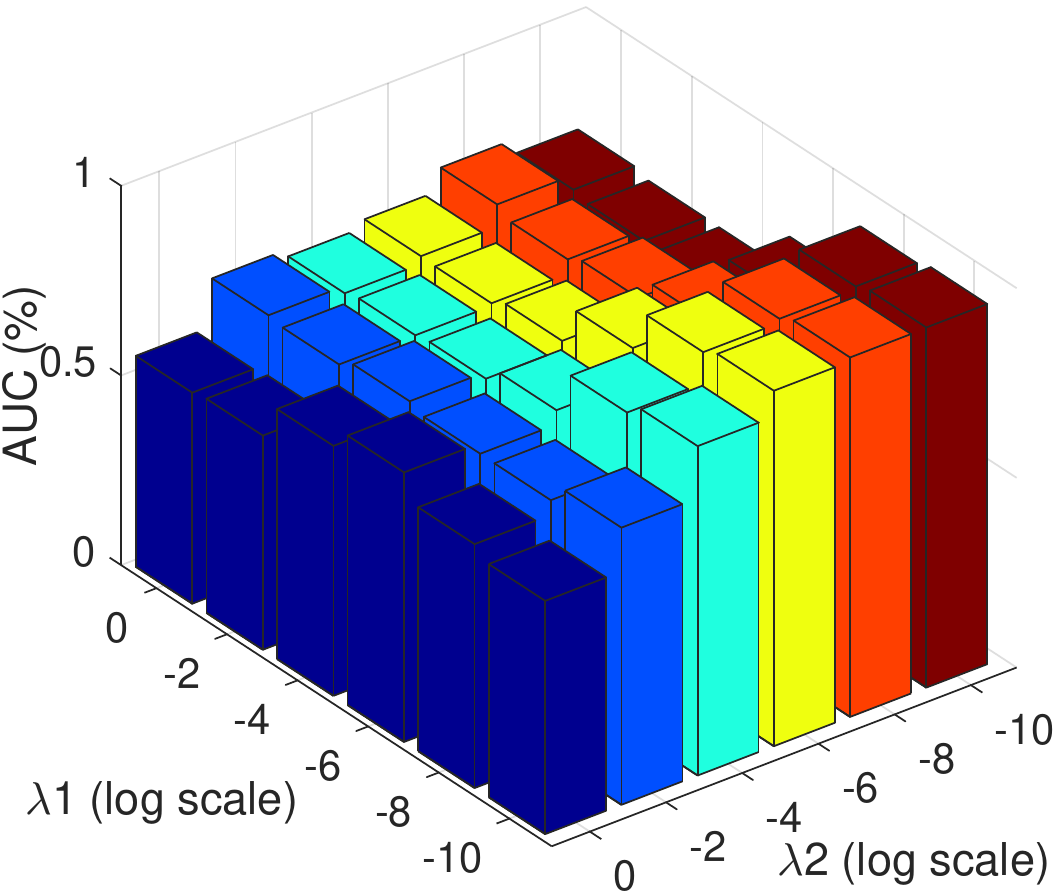}
        \end{minipage}}
        \vspace{-6.0mm}
         \caption{The effect of the regularization parameters $\lambda_1$ and $\lambda_2$ on Landmine (a) and Caltech-Birds (b) datasets, respectively.}
        \label{fig:Effect_Parameter_Classification} 
\end{figure}

\subsection{Effect of the Regularization Parameters $\lambda_1$ and $\lambda_2$}
To explore the effects of the regularization parameters $\lambda_1$ and $\lambda_2$ on Disjoint, School, Landmine and Caltech-Birds datasets, we fix other parameters and vary the parameters $\lambda_1$ and $\lambda_2$ in range $\{e^0, e^{-2},e^{-4},e^{-6},e^{-8},e^{-10}\}$. After randomly splitting each dataset into $50\%$ training and $50\%$ test set, we provide the results in Fig.~\ref{fig:Effect_Parameter_Regression} and Fig.~\ref{fig:Effect_Parameter_Classification}. From these experimental results, we can find that the corresponding regression and classification performances of our $\mr{FCL^3}$ model are stabile for most cases. From Fig.~\ref{fig:Effect_Parameter_Regression}(b), we can observe that the RMSE is much lower when the value of $\lambda_2$ is in the range from $e^{-2}$ to $e^{-4}$. For the other datasets, when $\lambda_1$ and $\lambda_2$ are both small, the model parameters can be very large, which can further increase the RMSE value and decrease the AUC value, respectively. This means that the assignment vector ${Z}^t$ is more important for both classification and regression problems. Meanwhile, this observation can also provide evidence that an appropriate model knowledge library can make the performance of our proposed $\mr{FCL^3}$ model better.

\begin{table}[htbp]
\caption{Comparisons of the effect of sparse autoencoder architecture on School and Landmine datasets, where $\mr{Ours\_woDecoder}$ denotes a unified framework without sparse autoencoder architecture.}
\vspace{-7.0pt}
\centering
\scalebox{0.89}{
\begin{tabular}{|c|c|c|c|c|c|}
\hline
 {}& Evaluation&CL3-Rand\cite{gan2018clusteredlifelong}&$\mr{Ours\_woDecoder} \atop \mr{-Rand}$ & Ours-Rand  \\
 \hline\hline
School        & RMSE&9.995$\pm$0.02     &9.990$\pm$0.02  &9.985$\pm$0.03       \\ \hline
Landmine     &AUC($\%$) & 78.904$\pm$0.47   &78.914$\pm $0.46  &79.022$\pm$0.42      \\   \hline
 \end{tabular}
}
\vspace{-10.0pt}
\label{table:effect_autoencoder}
\end{table}

\subsection{Effect of the Sparse Autoencoder}
Except for the effect of the regularization parameters $\lambda_1$ and $\lambda_2$, in this subsection, we explore how the sparse autoencoder architecture impacts on the performance of our model on School and Landmine datasets. To further emphasize the improvement on our conference paper, we also present the results of the CL3-Rand, which can be regarded as a two-phase framework without sparse autoencoder architecture, i.e., computing the assignment vector $Z^t$ and learning the new task. A unified framework without sparse autoencoder architecture is defined as $\mr{Ours\_woDecoder}$ in this experiment. As the results demonstrated in Table.~\ref{table:effect_autoencoder}, we can have the observation that 1) when comparing $\mr{Ours\_woDecoder}$ with CL3-Rand, a little improvement can be obtained, i.e., designing a unified lifelong framework can preserve both computing assignment vector $Z^t$ and learning the new task; 2) a similar trend can be observed across the sparse autoencoder architecture and it can be found that Ours-Rand achieves the best performance on both School and Landmine datasets. This observation confirms that the sparse autoencoder architecture assists in learning improved feature representation for the regression and classification tasks.

\renewcommand{\algorithmicrequire}{\textbf{Input:}}
\renewcommand{\algorithmicensure}{\textbf{Output:}}
\begin{algorithm}[t]
\caption{Optimizing Eq.~\eqref{eq:subproblem_z1} via Alternating Direction Method of Multipliers}
\begin{algorithmic}[1]
\REQUIRE Distance Vector $D\in\mb{R}^{K_t}$, $\lambda_2>0$, $\alpha>0$, $\beta>0$, Max-Iter;
\ENSURE $Z^t$; \\
\STATE Initialize ${Z}_0^{t}=J_0^t=I/(K_t+1)$, $d_0=-\gamma\mr{log}\Big(\frac{\min_{k}\left\|Ds_k-Ds_t\right\|_{\Omega_k^t}^2}{\sum_{k=1}^{K_t}\left\|Ds_k-Ds_t\right\|_{\Omega_k^t}^2}\Big)$ and ${D}=[D,d_0]$;
\WHILE {i<Max-Iter}
\STATE  Compute ${Z}_{i+1}^{t}$ via:\\
$\quad {Z}_{i+1}^{t}=\arg\min\limits_{Z}: \frac{\alpha}{\beta}\left\|Z\right\|_1+\frac{1}{2}\left\|Z-(J_i^t-\frac{U_{i}}{\beta})\right\|_F^2$;
\STATE  Compute $J_{i+1}^t$ via:
\begin{equation*}
\begin{aligned}
 J_{i+1}^t=\arg\min_{J}&: \frac{1}{2}\left\|J-(J_{i+1}+\frac{U_{i}+{D}}{\beta})\right\|_F^2;\\ s.t.&: 0\preceq \mr{vec}(J) \preceq \bm{1}_{K_t+1},\; J^{\top}\bm{1}_{K_t+1}=1;
 \end{aligned}
 \end{equation*}
\STATE Compute the Lagrange multiplier via: \\
$\quad U_{i+1}=U_{i}+\rho({Z}_{i+1}^t-J_{i+1}^t)$;
\IF {Convergence criteria satisfied}
\STATE Save $\{{Z}_{i+1}^t,J_{i+1}^t\}$;
\STATE Break;
\ENDIF
\STATE $i\leftarrow i+1$;
\ENDWHILE
\end{algorithmic}
\end{algorithm}

\vspace{-5pt}

\section{Conclusion} \label{sec:conclusion}
In this paper, we explore how to adapt flexible clustered lifelong machine learning system into a changing task environment, referred to as Flexible Clustered Lifelong Learning ($\mr{FCL^3}$) model. More specifically, our basic assumption is that all task models can be represented or described by multiple representative models, with each representative model corresponding to a task environment. We then propose to integrate representative models (i.e., model knowledge library) with discriminative feature learning (i.e., feature learning library), where model knowledge library can be self-selected via identifying and adding the new representative model, and feature learning library can learn a set of common discriminative representations among multiple tasks via a sparse autoencoder architecture. As a new task arrives, our $\mr{FCL^3}$ model can efficiently transfer knowledge from the representative models to learn the coming task via sparsity constraint, while redefining both model knowledge library and feature learning library when a higher outlier probability generates. After approximating each new task around mixture of representative models via Talyor expansion, and adopting alternating direction strategy solve lifelong machine learning subproblem, we conduct experiments on several real-world datasets; the presented experimental results demonstrate the efficiency and effectiveness of the proposed flexible clustered lifelong learning model. In the future, extending this work into deep networks will be our another attempt.


\appendices
\section{ADMM procedure for Eq.~\eqref{eq:subproblem_z1}}
This appendix gives the detail steps for Eq.~\eqref{eq:subproblem_z1}. In the following \textbf{Algorithm 3}, ${Z}_{i}^{t}$ denotes the $i$-th iteration of variable $Z^t$, and the solution of $J_{i}$ can be easily obtained by applying \emph{Karush-Kuhn-Tucker} (KKT) condition.


\ifCLASSOPTIONcompsoc
  \section*{Acknowledgments}
\else
  \section*{Acknowledgment}
\fi

The authors would like to thank Prof. Eric Eaton for his constructive suggestion. The research was supported by National Natural Science Foundation of China under Grant (61722311, U1613214, 61821005, 61533015) and LiaoNing Revitalization Talents Program (XLYC1807053).

\ifCLASSOPTIONcaptionsoff
  \newpage
\fi



%

\vspace{-10pt}
\bibliographystyle{plain}
\bibliography{Multitask}

\begin{IEEEbiography}[{\includegraphics[width=1.0in,height=1.26in,clip,keepaspectratio]{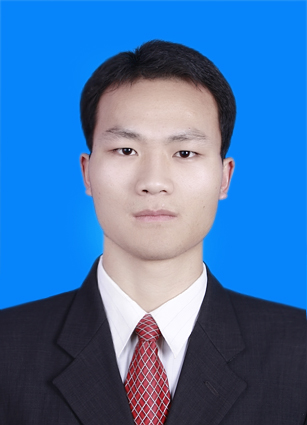}}]{Gan Sun} Gan Sun is currently a Ph. D candidate in State Key Laboratory of Robotics, Shenyang Institute of Automation, University of Chinese Academy of Sciences. He received the B.S. degree from Shandong Agricultural University in 2013. His current research interests include lifelong learning, multi-task learning, metric learning, online parallel learning, sparse learning and deep learning.
\end{IEEEbiography}

\vspace{-45pt}

\begin{IEEEbiography}[{\includegraphics[width=1in,height=1.25in,clip,keepaspectratio]{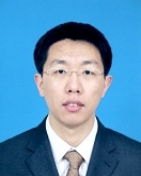}}]{Yang Cong} Yang Cong (S’09-M’11-SM’15) is a full professor of Chinese Academy of Sciences. He received the B.Sc. degree from Northeast University in 2004, and the Ph.D. degree from State Key Laboratory of Robotics, Chinese Academy of Sciences in 2009. He was a Research Fellow of National University of Singapore (NUS) and Nanyang Technological University (NTU) from 2009 to 2011, respectively; and a visiting scholar of University of Rochester. He has served on the editorial board of the Journal of Multimedia. His current research interests include image processing, compute vision, machine learning, multimedia, medical imaging, data mining and robot navigation. He has authored over 70 technical papers. He is also a senior member of IEEE.
\end{IEEEbiography}

\vspace{-50pt}

\begin{IEEEbiography}
[{\includegraphics[width=1in,height=1.25in, clip, keepaspectratio]{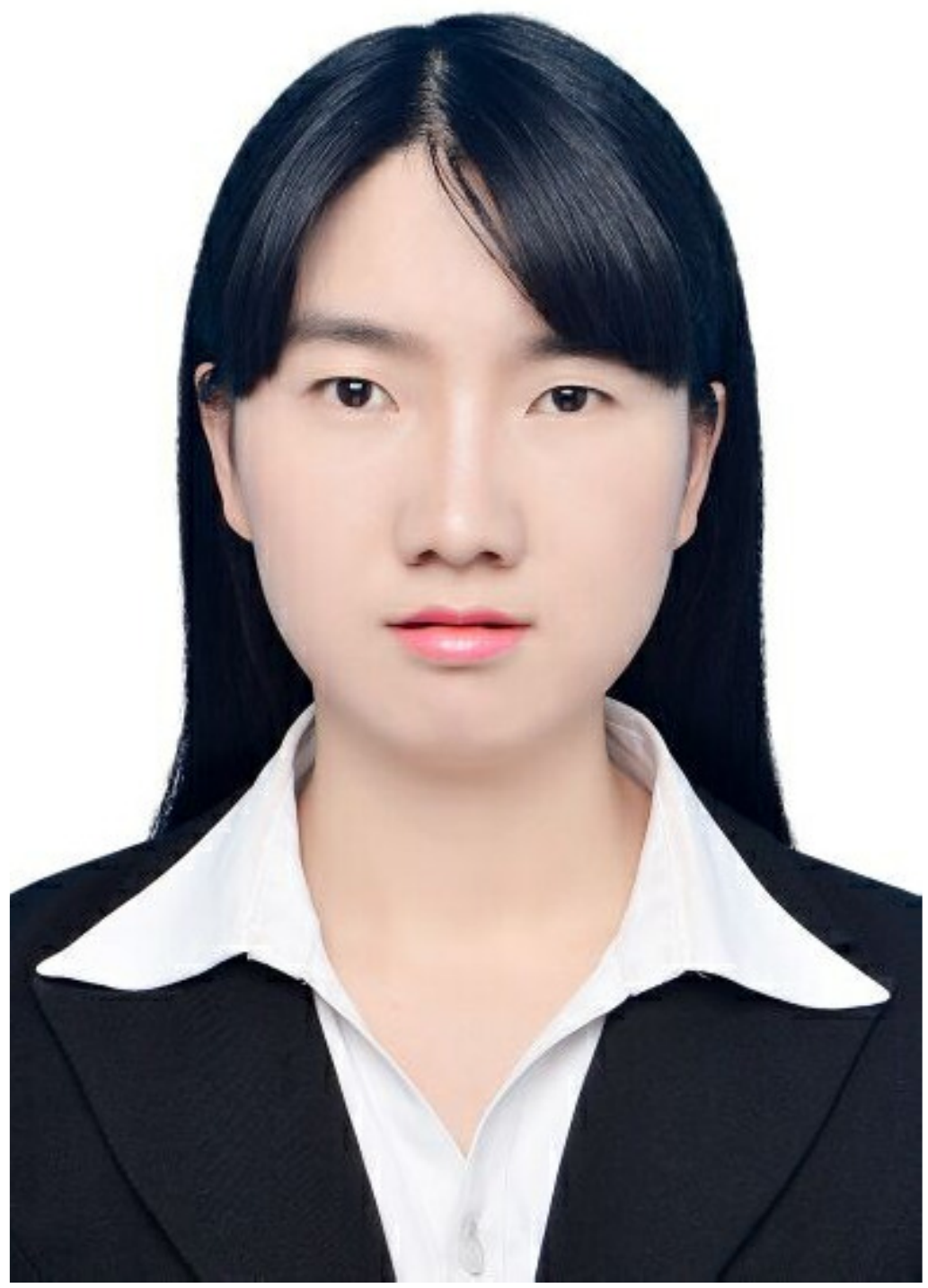}}]{Qianqian Wang}
Qianqian Wang received the B.Eng. degree from Lanzhou University of Technology, Lanzhou, China, in 2014. Now, she is working toward the Ph.D. degree at Xidian University, Xi'an, China. Her research interests include dimensionality reduction, pattern recognition and deep learning.
\end{IEEEbiography}

\vspace{-50pt}

\begin{IEEEbiography}
[{\includegraphics[width=1in,height=1.25in, clip, keepaspectratio]{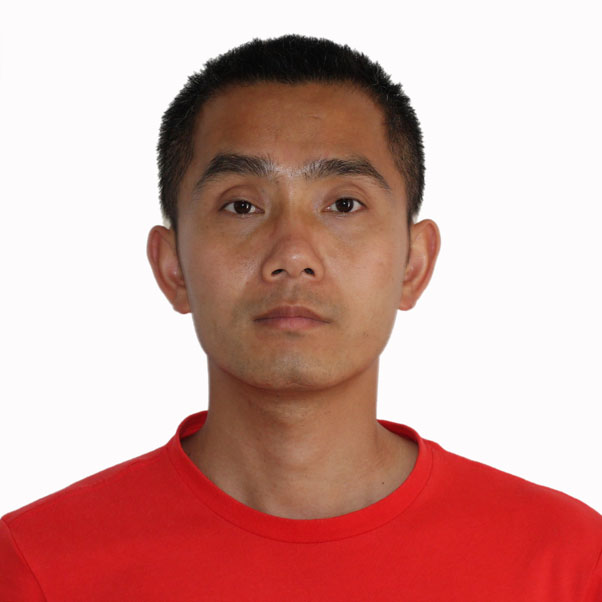}}]{Bineng Zhong}
Bineng Zhong received the B.S., M.S., and Ph.D. degrees in computer science from the Harbin Institute of Technology, Harbin, China, in 2004, 2006, and 2010, respectively. From 2007 to 2008, he was a Research Fellow with the Institute of Automation and Institute of Computing Technology, Chinese Academy of Science. From September 2017 to September 2018, he is a visiting scholar in Northeastern University, Boston, MA, USA. Currently, he is an professor with the School of Computer Science and Technology, Huaqiao University, Xiamen, China. His current research interests include pattern recognition, machine learning, and computer vision.
\end{IEEEbiography}

\vspace{-50pt}

\begin{IEEEbiography}
[{\includegraphics[width=1in,height=1.29in, clip, keepaspectratio]{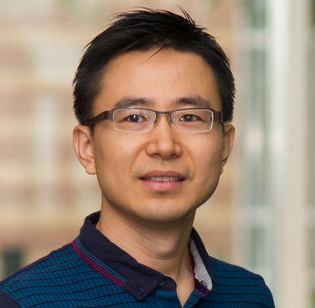}}]{Yun Fu}
Yun Fu (S'07-M'08-SM'11-F'19) received the B.Eng. degree in information engineering and the M.Eng. degree in pattern recognition and intelligence systems from Xi'an Jiaotong University, China, respectively, and the M.S. degree in statistics and the Ph.D. degree in electrical and computer engineering from the University of Illinois at Urbana-Champaign, respectively. He is an interdisciplinary faculty member affiliated with College of Engineering and the Khoury College of Computer and Information Sciences at Northeastern University since 2012. His research interests are Machine Learning, Computational Intelligence, Big Data Mining, Computer Vision, Pattern Recognition, and Cyber-Physical Systems. He has extensive publications in leading journals, books/book chapters and international conferences/workshops. He serves as associate editor, chairs, PC member and reviewer of many top journals and international conferences/workshops. He received seven Prestigious Young Investigator Awards from NAE, ONR, ARO, IEEE, INNS, UIUC, Grainger Foundation; nine Best Paper Awards from IEEE, IAPR, SPIE, SIAM; many major Industrial Research Awards from Google, Samsung, and Adobe, etc. He is currently an Associate Editor of the IEEE Transactions on Neural Networks and Leaning Systems (TNNLS). He is fellow of IEEE, IAPR, OSA and SPIE, a Lifetime Distinguished Member of ACM, Lifetime Member of AAAI and Institute of Mathematical Statistics, member of ACM Future of Computing Academy, Global Young Academy, AAAS, INNS and Beckman Graduate Fellow during 2007-2008.
\end{IEEEbiography}


\end{document}